# A Visual Big Data System for the Prediction of Weather-related Variables: Jordan-Spain Case Study

Shadi Aljawarneh[1], Juan A. Lara[2,*], Muneer Bani Yassein[1]

[1]Jordan University of Science and Technology, JUST, Faculty of Computer and Information Technology, P.O.Box 3030, 22110, Irbid, Jordan.

[2]Madrid Open University, UDIMA, School of Computer Science, Carretera de La Coruña, KM. 38,500, Vía de Servicio, nº 15, 28400 Collado Villalba, Madrid, Spain. *Correspondence: juanalfonso.lara@udima.es

**Abstract**: The Meteorology is a field where huge amounts of data are generated, mainly collected by sensors at weather stations, where different variables can be measured. Those data have some particularities such as high volume and dimensionality, the frequent existence of missing values in some stations, and the high correlation between collected variables. In this regard, it is crucial to make use of Big Data and Data Mining techniques to deal with those data and extract useful knowledge from them that can be used, for instance, to predict weather phenomena. In this paper, we propose a visual big data system that is designed to deal with high amounts of weather-related data and lets the user analyze those data to perform predictive tasks over the considered variables (temperature and rainfall). The proposed system collects open data and loads them onto a local NoSQL database fusing them at different levels of temporal and spatial aggregation in order to perform a predictive analysis using univariate and multivariate approaches as well as forecasting based on training data from neighbor stations in cases with high rates of missing values. The system has been assessed in terms of usability and predictive performance, obtaining an overall normalized mean squared error value of 0.00013, and an overall directional symmetry value of nearly 0.84. Our system has been rated positively by a group of experts in the area (all aspects of the system except graphic desing were rated 3 or above in a 1-5 scale). The promising preliminary results obtained demonstrate the validity of our system and invite us to keep working on this area.

**Keywords**: Big Data; Weather forecasting; Data Mining; Information Fusion; MongoDB.

## 1. Introduction

Big Data is a term that refers to huge and complex amounts of data that needs the application of computing techniques to be managed. By extension, it also refers to a series of procedures and methods that are used to capture, store, visualize, and analyze those amounts of data in search of interesting and value patterns [1]. Most authors agree that data should meet a series of requirements so that they can be called big data. Those are the so-called three V's (Volume, Variety, and Velocity). Other authors have more recently added other V's such as Veracity or Variability, to name a few [2].

Big Data may come from a series of environments such as sensor networks [3], smart cities [4], social networks [5], medical devices [6], or transactional procedures [7], to name a few examples. Those data need to be stored and NoSQL (Not only SQL -Structure Query Language-) databases are being used quite commonly [8]. After that, data are prepared and analyzed using approaches such as the Knowledge Discovery in Databases (KDD) process [9], based on the application of Data Mining techniques in search of useful knowledge.

Big Data has demonstrated to be useful in different fields such as medicine [10], sport [11], stock market [12], and many other fields that we could imagine as long as enough amounts of data are generated.

In addition to those cases, one of the areas where Big Data has been widely used in the last few years is Meteorology [13]. It is an area where enormous amounts of data are generated. For example, the sensors located on weather stations, public and private, are an important source of data. Those sensors can collect information about different variables, such as temperature, rainfalls, humidity, wind speed and direction, and so on. The weather-related data has some important characteristics that make it particularly interesting to analyze them:

a) We can find huge amounts of data in terms of all dimensions (V's) of Big Data.

b) It is quite frequent to find lots of missing values in certain weather stations since measuring infrastructure needs improvement in some local areas.

c) The measured variables are usually correlated.

Those features make it necessary for the application of Big Data and Data Mining techniques in order to properly manage weather data in search of knowledge. In particular, predictive data mining techniques have been widely applied in the last decades because one of the main purposes of weather data analysis, if not the most important, is forecasting [14].

In this paper, we present a visual system that can deal with weather-related big data. In particular, our system collects data from an open repository and loads them onto a local NoSQL database fusing them at different levels of temporal and spatial aggregation. The user can visualize data and prepare them for analysis by using the different multimedia resources of the system, mainly text, graphics, and maps. In this regard, predictive time series techniques are applied, both with univariate (one dimension) and multivariate (several dimensions) analysis approaches. Besides, the system lets the user deal with weather stations with lots of missing value by predicting through interpolation with data from other neighbor stations with more complete data. A system like this can be useful in weather decision-making since it may help in areas such as hydrological resources management or crop sowing planning.

Our system has been validated in terms of usability, by using a user-center design approach; and predictive performance, as well as using a panel of experts. The results obtained demonstrate that our system is well rated by users and experts in terms of its ability to perform the tasks it has been designed for, usefulness, ease of use, and recommendation of use. The results also prove the predictive ability of our system measured in terms of mean square error.

The purpose of this paper is to present the implemented system, which can deal with weather big data and make accurate forecasting of different meteorological phenomena. This paper mainly contributes to the area by presenting a system that can help other experts who intend to develop similar projects in which particularities such as big data, missing values, and highly correlated variables may appear. In this regard, this paper may be a reference for other authors who can apply similar ideas, techniques, or tools as the ones presented in this paper.

The remainder of the paper is organized as follows. Section 2 gives an overview of the major big data and predictive data mining techniques and presents the main related works with regards to weather-related variables forecasting. Section 3 describes the proposed big data system. Section 4 presents the validation of the system in terms of usability, experts' judgment, and predictive power. In Section 5 we discuss the proposed system and the results obtained. Finally, Section 6 outlines the conclusions and future work concerning the proposed system.

## 2. Background

As we have already stated in the Introduction section, the term Big Data refers to a series of techniques aiming at dealing with huge amounts of data in search of some kind of useful knowledge. There are different tasks to carry out when it comes to the development of a Big Data project.

One first key aspect is the storage of data. In this regard, NoSQL databases have demonstrated to be one of the most used solutions in big data projects [15]. In these kinds of databases, the stored data do not need a defined structure such as occurs with the tables of relational models. NoSQL systems do not use SQL for querying purposes. On the contrary, data may be stored using other types of approaches and we could split NoSQL databases into four different categories: document-oriented, key-value, wide column, and graph-oriented [15]. There are different NoSQL systems and some of the most popular are Infinispan [16], Cassandra [17], InfiniteGraph [18], or MongoDB [19], to name a few. In this research, we have used MongoDB, which is a document-oriented NoSQL system where data are stored according to BSON (Binary JSON -JavaScript Object Notation-) specification [20]. Documents are grouped into collections that belong to a particular database. Likewise other NoSQL systems, MongoDB has limitations. However, it also has several advantages that led us to use it. In particular, it is a free and open-source multi-platform system, it provides high performance and automatic scalability, it permits to organize documents easily, and it is easy to integrate with software applications for most languages [21]. In addition, we also decided to use this platform due to our previous positive experience in similar academic projects.

Regarding the rest of the tasks, such as data preparation, data analysis, and model evaluation, we followed the so-called KDD Process [9], which pursues the extraction of useful, previously unknown, non-trivial, and implicit knowledge through a series of phases. In our project, the focus is set on the Data Mining phase, which aims at extracting useful knowledge from prepared data (minable view) by addressing a series of problems. Particularly, we have conducted predictive tasks [22] by addressing the problems of classification [23] and regression [24]. Those are problems in which the purpose is to predict an unknown value for a certain instance (in our case, the temperatures and rainfall for each month of the next year). There are many different types of predictive data mining techniques and we have used a variety of them in this project, such as Gaussian Processes (GP) [25], Multilayer Perceptron (MLP, a type of Neural Network) [26], SMOreg (SMO, a Support Vector Machine-SVM) [27] and Linear Regression (LR) [28]. We decided to use them because these types of techniques have demonstrated to be useful in the big data landscape [25] [29] [30]. In particular, we have used the implementation of those techniques provided by Weka [31], which is a tool for Data Mining easy to integrate with applications due to its powerful API (Application Programming Interface). It is important to clarify that we have applied those techniques on time series data, whose treatment, in general, has a series of particularities and implications [32]. The use of time-series data is quite common in Weather-related knowledge extraction projects and some study cases can be found in [33] [34].

Continuing with the KDD process, we have also taken some ideas from it concerning the evaluation of the predictive models obtained. The KDD process suggests the development of an Evaluation stage in which models need to be assessed to determine how accurate or powerful they are. In our case, we need to evaluate how the predictions made differ from the actual values. In this regard, the use of indicators such as MAE (Mean Absolute Error) or MSE (Mean Squared Error) has demonstrated to be useful in big data projects [35].

In this section, it is also important to consider the most representative related works that have applied Big Data Mining techniques to the area of Meteorology in the last few years [36]. There is a group of papers that present some preliminary implementations. In [37] [38] the authors present an analytical Big Data prediction framework for temperature forecasting with monthly granularity, based on Hadoop and MapReduce algorithm. This paper seems to be focused on how data are stored and managed and it does not include a comprehensive qualitative assessment of the forecasting ability of their approach. In [39] the authors present an interesting case study about solar power prediction but again no details are provided about the validation of the predictive power of the system.

Other works are mainly focused on computational issues. In [40] the authors work on data analysis using Apache Hadoop and Apache Spark. They perform experiments to decide the best tools among Hadoop using Pig and Hive queries. In [41] the authors present a report on the real-world application of global atmospheric ensemble Data Assimilation with a large sample size, focusing on variables such as atmospheric (not Earth surface) temperature and humidity.

We can also find papers interested in the generation of features for the weather forecast. In [42] the authors employ auto-encoders to simulate hourly weather data in 30 years and can automatically learn the features from a massive volume of data via layer-by-layer feature granulation. The results obtained by the authors show the power of the generated features in improving the predictive accuracy in time series problems.

Finally, more mature linked works can be found in the literature. In [43] the authors propose a deep learning-based weather forecast system and conduct data volume and recency analysis. This work focuses on one particular weather station located in Baltimore Washington International Airport, and forecast the value of temperature in that station. The predictive error is measured using the MAE indicator. However, the topic of this paper is not the prediction itself but the discovery of relationships with prediction accuracy and data volume/recency. In [44], they apply the concept on a dynamic data-driven system to obtain correlations between the prediction goals and several different combination results, by employing association analysis, sequence analysis, and stacked auto-encoders. They intend to predict the variable Temperature with hourly granularity and evaluate their system by using the Normalized MSE indicator.

To conclude this section, it is convenient to state that, to the best of our knowledge, our system is the first one that provides the user with visual and multimedia functionalities for predicting several Earth surface weather-related variables in all regions of the world with a monthly granularity by using information fusion approaches and makes an effort to improve predictive accuracy by addressing the problem of the existing high rate of missing values in some stations and taking advantage of the correlation between those variables. Also, the authors did not find any other similar system developed by utilizing user-center design techniques and validated from both qualitative and quantitative approaches.

## 3. System description

In this section, we will describe the system from two perspectives. First, we will explain the architecture of the system and provide some preliminary details about its key components, their interactions, and the main flows of information (section 3.1). Second, we will provide more detailed information about each part of the system and describe the main tasks performed by the system (3.2). Note that just a summarized description of the system is presented here and more details about its implementation can be found in [45] [46].

### 3.1. System working

The proposed system works in a way that it takes data from an open repository on the Internet and saves them onto a local database. After that, the user can create particular views of that local database in order to focus on certain geographical regions. The user can visualize data and create minable views to perform a predictive analysis based on it. Finally, the system shows the results of the prediction performed. This process can be summarized from the perspective of components in Figure 1.

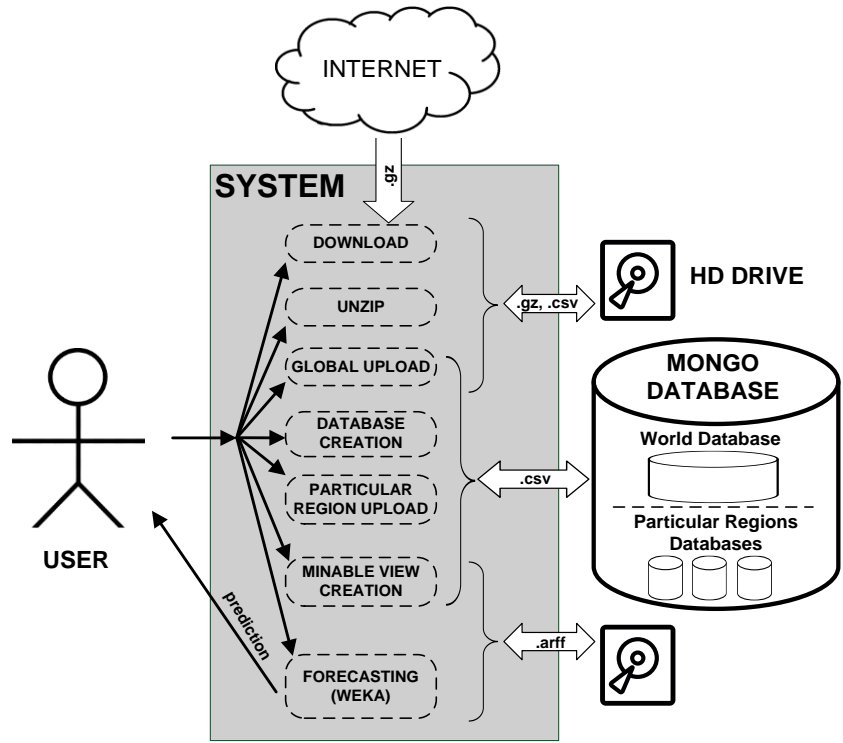

**Figure 1**. Main components of the system.

As we can see, it is the user who starts each of the main functionalities provided by the system, than can be invoked to download data from the Internet repository, unzip those data, load data onto a global world database, create a particular database for a certain region of the world, load that database with data, create a minable view for analysis and, finally, forecast the desired variable. We notice that data are downloaded from the Internet in a compressed format, particularly .gz, and then the corresponding files are unzipped and left on the local hard disk (HD) drive as .csv files that are used to feed the global world database. There is a group of functionalities that let the user create particular geographical regions databases and feed them with .csv data; then, the loaded data can be selected and read in order to create minable views in the form of .arff files that are used by the forecasting module, which provides the user with the predicted value of the selected variable. In Figure 1, note that the HD drive has been represented twice for the sake of readability but it represents the same drive indeed.

If we consider a view of the system focused on information flow, we could present it as depicted in Figure 2. The process begins when the users download from the Internet data associated with a particular year and this step is repeated for as many years as he or she wishes. Then the downloaded files are unzipped and data are loaded onto the global world database. After that, the user can create a database for a particular region of the world and will be loaded with data of the desired years, as long as they have previously loaded onto the global database. Next, the user can visualize data and then create a minable view for analysis. He or she will open that minable view and may use it for performing as many predictions as the system is designed for (to be discussed later). The process will end up once the user decides not to perform further predictions.

It is important to clarify that the flow depicted in Figure 2 represents the standard and most common course of action that the user is expected to carry out. However, the user could go back to some previous stage at any time. This way, the process could be said to be iterative and sequential, similarly to the KDD process on which it is inspired.

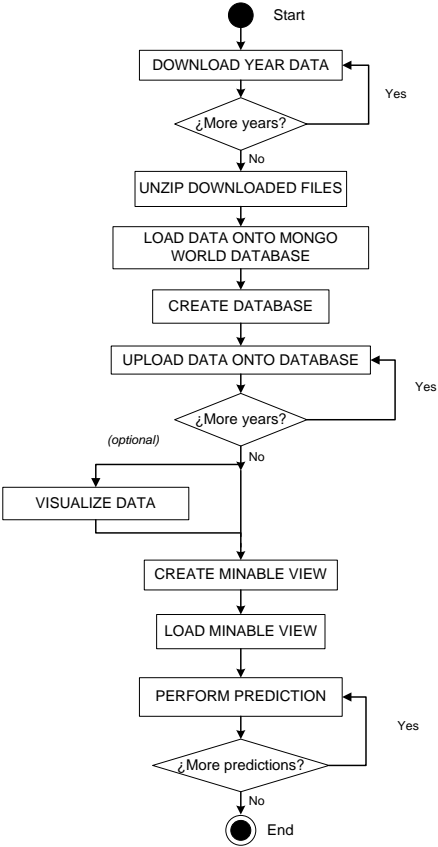

**Figure 2**. Flowchart depicting the behavior of the system.

**3.2. Implementation and functional description of the system**

Next, we will describe in detail the main components of the system and provide some screenshots of the system interface and system output in order to illustrate how the system works. In this case, we will provide a textual description of the system's components. Before, it is important t to clarify that all the components have been implemented using the language Java, NetBeans IDE 8.2 (Patch 2), and Java SE Runtime Environment 1.8.0_191-b12. The graphical part of our single-windows application has been implemented using AWT and the library Swing. We used the Weka package for Java applications. Finally, we used MongoDB package for Java applications and the tools MongoDB Compass 1.19.12 and Studio 3T 2020.2 as MongoDB graphical interfaces. In particular, we used them to carry out a pre-load of data onto our database in which we included some important information about geographical regions, countries, and weather stations inside each country. Note that we have not used a cluster of computers to run our system, but we did it on the localhost for it being a prototype.

A. DOWNLOAD

The purpose of this component of the system is to download the data from the Internet. To do so, the user introduces the year for which data will be downloaded, one year at a time (left side of Figure 3). After that, the user will press on "Download button" and the download process will begin. A progress bar has been implemented so that the user knows the percentage of completion of this task.

Once the user presses the "Download button" the system connects to the public repository containing the data. In particular, data are obtained from the repository GHCN† (Global Historical Climatology Network), which integrates daily climate sources of data recorded in stations all over the Earth. It is an incremental and dynamical repository that is continuously assessed in terms of quality control. The repository contains historical information that dates back from the mid-eighteenth century to now. This repository is managed by the National Oceanic and Atmospheric Administration (NOAA). Note that the used repository store information from more than 100.000 weather stations from around 180 countries. The measured variables are: daily minimum temperature (TMIN), daily maximum temperature (TMAX), the temperature at the time of observation (TOBS), precipitation such as rain or melted snow (PRCP), snowfall (SNOW) and snow depth (SNWD) are recorded.

In order to download the data of a particular year, the system builds the next URL, substituting the string "yyyy" by the intended year: **https://www1.ncdc.noaa.gov/pub/data/ghcn/daily/by_year/yyyy.csv.gz**. As we can see, data are downloaded as zipped .csv.gz files that are stored on the local hard disk drive at the location (folder) indicated by the user. The size of each of those files is around 200 MegaBytes and takes a few minutes for downloading depending on the connection.

Note that the component in charge of implementing this functionality has been implemented as a thread (Runnable in Java) and makes use of packages java.io, java.net, java.util, and javax.swing.

B. UNZIP

Once the .gz files are stored in the disk drive, the next step is to unzip them, one file at a time. The user will click on the "Unzip" button (middle of Figure 3) and a window will open to let him/her select a .gz file from the local hard disk drive. Afterward, the system will unzip the selected file and the resulting unzipped file (.csv) will be stored at the location (folder) indicated by the user on the local disk drive as well. There is a bar showing the progress of this task that will take from a few seconds to one minute depending on the computer and the HD drive features. The compression rate of .gz files is around 1/6, which means that a .gz file of, for instance, 200 MegaBytes will occupy around 1.2 GigaBytes when unzipped as a .csv file. The structure of .csv files consists of eight fields separated by commas that appear in each row of the .csv file:

- Station ID: 11 characters encoding the weather station (e.g., CA007020860).
- Date: 8 characters representing the data of recording, with the format YYYYMMDD.

---

† https://www.ncdc.noaa.gov/data-access/land-based-station-data/land-based-datasets/global-historical-climatology-network-ghcn

- Measured Variable: 4 characters that encode the variable collected (TMIN, TMAX, TOBS, PRCP, SNOW, SNWD).

- Recorded Value: Integer value that represents the value for the measured variable. The units used for each variable are: TMIN, TMAX, and TOBS (tenths of ºC); PRCP (tenths of mm); SNOW (mm); and SNWD (mm).

- M-Flag: 1 character - Measurement flag[‡].

- Q-Flag: 1 character - Quality flag.

- S-Flag: 1 character - Source flag.

- Observation time: 4 characters indicating the time of recording (for instance, "0700" represents "07:00").

In our research, we have worked with the variables TMIN, TMAX, and PRCP. Note that the component in charge of implementing this functionality makes use of packages java.io, java.util, and javax.swing.

C. GLOBAL UPLOAD

After unzipping the file, the next step will be to take and load the data onto a MongoDB database. The user will click on the button labeled "Export CSV to MongoDB" (right side of Figure 3) and then the system will open a new window that will let the user select the .csv file to read. After selecting that file, the system will load data contained in the .csv file selected by the user onto the database. There is a progress bar showing the status of the process, which will be repeated for each .csv file that is intended to be loaded by the user.

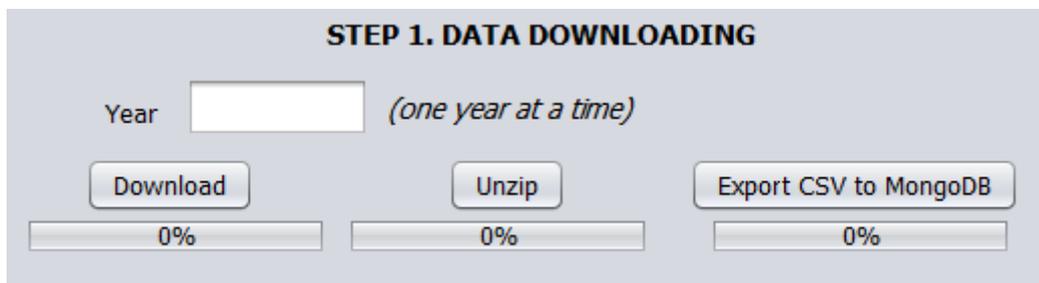

**Figure 3**. Download Unzip and Global Upload functionalities interface.

It is important to clarify that the MongoDB database used in this step is composed of a series of collections, one per year. For instance, the collection named "2017" will contain the date relative to that year for all the stations of the world. Therefore, we could say it is a worldwide database, labeled as "World Database" in Figure 1.

Each collection in this database (e.g., "2017") is composed of several documents equal to the number of rows contained in the .csv loaded file. That is, there is a document for each recorded value at a certain station for a certain variable at a certain observation time. Each document is composed of the following fields: MongoDB object ID, date, variable (TMIN, TMAX, or PRCP), value, and observation time. The data load is carried out by the corresponding module by executing this MongoDB command:

---

[‡] Check ftp://ftp.ncdc.noaa.gov/pub/data/ghcn/daily/readme.txt for further information on the M-, Q-, and S-Flag.

```
mongoimport --db <database_name> --collection <collection_name> --type csv –file
<file_route>\<file_name>.csv --fields StationID, date, variable, value, observ_time
```

For each collection (year) a number of about 35 million documents are created, which represents around 5 GigaBytes of data (per year) in the hard disk. No wonder the load process takes a long time (a good number of minutes in some cases).

Note that the component in charge of implementing this functionality has been implemented as a thread (Runnable in Java) and makes use of packages java.util, javax.swing, com.mongodb, and org.bson.

D. DATABASE CREATION

The purpose of this step is to create a particular database where we can load data from a particular area on which we intend to focus for later analysis. In this regard, there are three possibilities. The user may create a database for: a) a particular weather station; b) a particular country; or b) a wider geographical region, such as Europe, North America, etc. The user will just select the desired station/country/region and click on the corresponding button (see Figure 4).

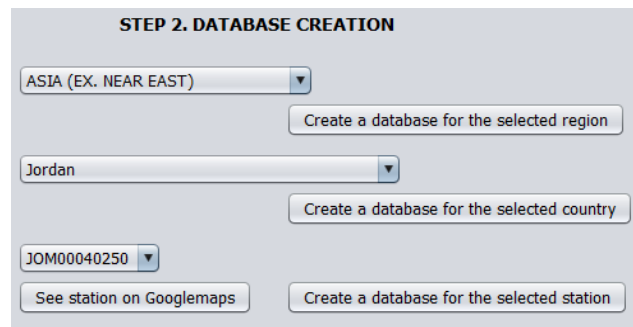

**Figure 4**. Database creation functionality interface.

After that, a new database will be created (denoted as "Particular Regions Databases" in Figure 1) in our MongoDB system. The new database will take the name of the corresponding station, country, or region. When created, the new database will only contain 3 empty collections inside: total_p, total_tmax, and total_tmin. As we will see in the next component, those collections intend to store the aggregated data for variables precipitation, maximum temperature, and minimum temperature, respectively. This is a quick step (just a few seconds) since only a few simple transactions have to be performed on the database.

We have implemented functionality that lets the user see on a Google map the particular location of a certain station. To do so, the user will click on the button "See station on GoogleMaps" and a new window will open on the web browser containing a Google map that shows the location of the currently selected station (see Figure 5 for an example). It is important to clarify that, before the first run of the system, we have pre-loaded a database in which we have stored a list of all the regions, countries, and stations (3 different collections in MongoDB). In particular, for each region we store its name; for each country, we store its name and region; and for each station, we store information about its country, region, latitude, and longitude[§].

---

[§] All that information has been obtained from https://www1.ncdc.noaa.gov/pub/data/ghcn/daily/ghcnd-stations.txt

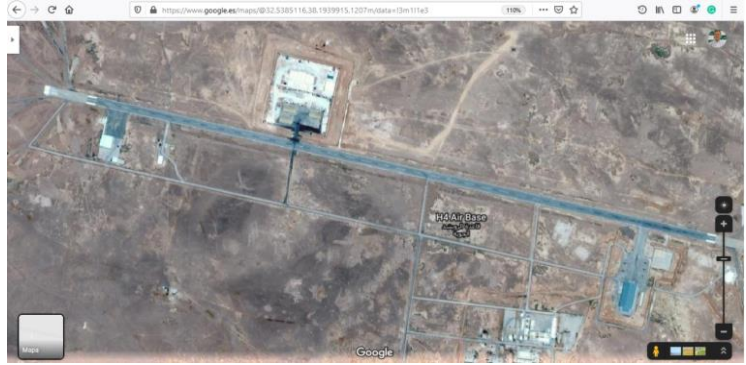

**Figure 5**. Web browser window showing a GoogleMap with the location of the station "JOM00040250".

Note that the component in charge of implementing this functionality makes use of packages java.util, javax.swing, and com.mongodb.

E. PARTICULAR REGION UPLOAD

After creating a particular database for analysis, we need to load data on it. To do so, the user will be able to select the database previously created, the year whose data will be loaded onto the created database, and click on the button "Update database" (see Figure 6). This process will be repeated, once database and year at a time, and will be only available for databases created by module "D. DATABASE CREATION" and for years' data loaded on the worldwide database by module "C. GLOBAL UPLOAD". There is a progress bar so that the user may know the status of the process.

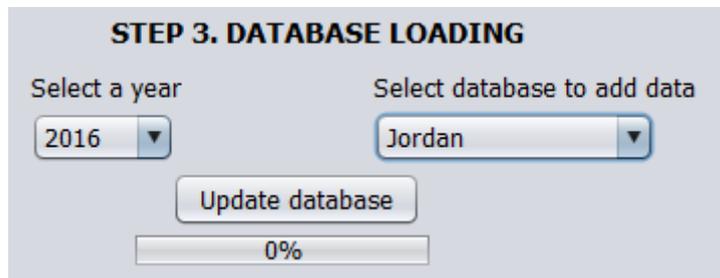

**Figure 6**. Particular Region Upload functionality interface.

This process involves many transactions on the worldwide database (from where data are read) and the selected database (where data are written), so it takes a long time, usually some minutes, depending on the computer and hard disk drive.

First, three new collections will be created. For instance, if the selected database and year were "Jordan" and "2017" respectively, inside "Jordan" database the system will create the three new collections named "2017", "2017TMAX" and "2017TMIN", containing, respectively, information about precipitation, maximum temperature and minimum temperature for every single day (aggregated by day) and for each station (if it is a "station database", then we will have only one station; if it is a "country" or "region" database, then we will have all the stations geographically contained in that country or region). Each of those collections will contain different documents with the following information: station ID, date, and recorded value (see Figure 7). In particular, each collection will contain as many documents as different stations and days of the year we have in the database. Therefore, we have a temporal aggregation at this point since all data for each particular day are aggregated. As it is a temporal aggregation it makes sense to aggregate values by using the "addition" operator for precipitation, the "minimum" operator for minimum temperature, and the "maximum" operator for maximum temperature. Note that those documents contain data aggregated on a daily basis and therefore the storage size reduces significantly (a few dozens KiloBytes per collection).

Second, the collections created in step "D. DATABASE CREATION" (total_p, total_tmax, and total_tmin) are updated. Particularly, twelve new documents will be added to each of those 3 collections, each containing the aggregated values for each month of the selected year considering all the stations of the selected database. In this case, there are two kinds of aggregations, temporal and spatial. For temporal aggregation, we have followed the same aggregation approach defined in the previous paragraph with the only difference that here we have a higher level of temporal aggregation: monthly. For spatial aggregation, as we need to fuse data from different stations, we decided to use the "average" operator for all variables. If the value cannot be calculated due to an excessive rate of missing values, a "?" symbol will be stored. Here, the aggregation level is even higher so the storage size decreases considerably (a low number of KiloBytes per collection). In these aggregated collections, which will be the base in our analysis module, we have stored data using some new units that will be utilized during the later mining process. In particular, for maximum and minimum temperature, we have used both Celsius and Fahrenheit degrees (the former one was just tenths of Celsius), while "mm" is the new unit for precipitation (the former one was tenths of mm). Note that the necessary basic conversions have been performed from the worldwide database data to obtain the values according to the new units. Besides, we have also lightly changed the date format from "yyyymmdd" into ISOdate format "yyyy-mm-ddThh:mm:ss.mmmZ", in search of easier integration with analysis modules described later.

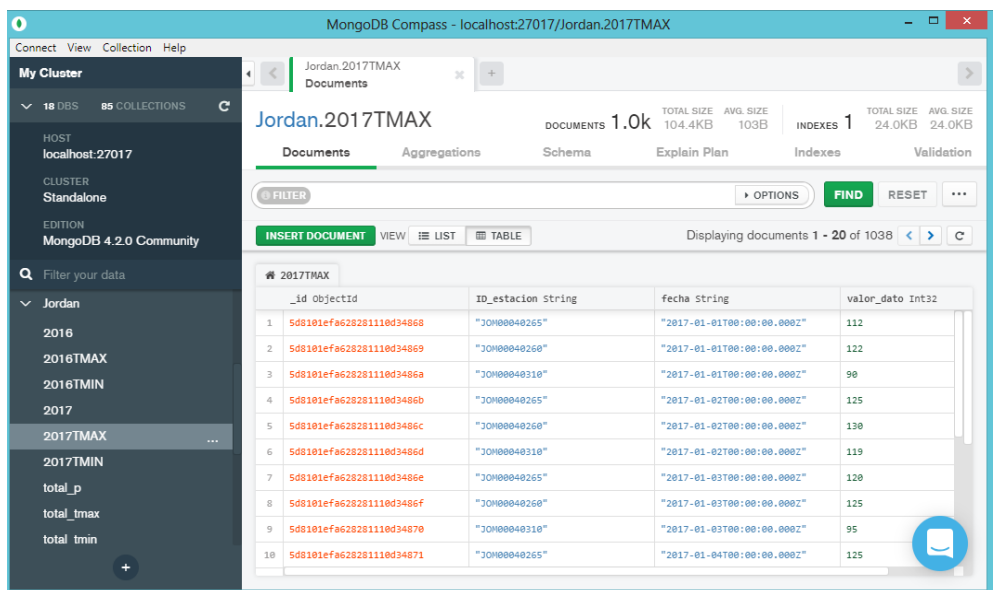

**Figure 7**. Excerpt from MongoDB database for a particular country database[**].

Note that the component in charge of implementing this functionality has been implemented as a thread (Runnable in Java) and makes use of packages java.util, javax.swing, com.mongodb, and org.bson. Particularly, those last two packages provide the programmer with a powerful API for reading and writing on MongoDB databases, preventing the programmer from having to write tough native MongoDB commands.

---

[**] There are some Spanish word in the figure whose meaning is: estación = station; fecha = date ; valor_dato = datum_value

F.  MINABLE VIEW CREATION

The purpose of this step is to create a minable view of data that can be used for forecasting purposes, as we will see in the next module. Before that, the user may use the functionality to visualize data so he or she can have a better understanding of the data to be used. To do so, the user will select the range of years to visualize, the variable, and the database, before selecting the option "Show on diagram" and clicking on the button "Send" (see Figure 9). After that, the system will display the corresponding 2-dimensions (x=time; y=weather variable) diagram in the form of a time series (see Figure 8) and will inform the user about the rate of the missing values of the displayed data. The visualization of data is an important functionality before building a minable view.

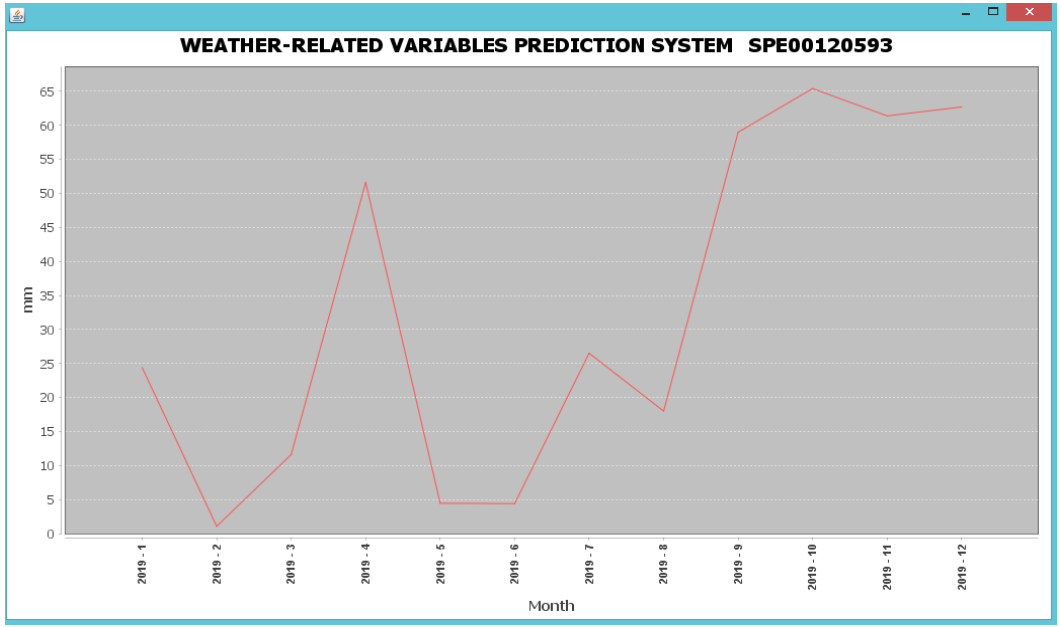

**Figure 8**. Example of a diagram that users may visualize on the system.

Before explaining how minable views are created, it is crucial to explain that our system lets the user perform two kinds of analysis, called "standard" and "neighbour-based" analysis.

In the **"standard" analysis**, the user will select the range of years (From-To in Figure 9) to consider for creating the minable view, and the database. After that, the user will select the option "Create .ARFF for standard analysis" and click on the button "Send", as depicted in Figure 9 (note that the field "Variable" of this part of the interface is not relevant in this type of analysis since data from all variables will be used to build the minable view).

After that, the system read from MongoDB the data for the selected database and years and will create a file with an extension .ARFF similar to the one shown in Appendix I.A. It is a kind of comma-separated format file that will be stored in the location of the hard disk drive indicated by the user. In our case, 4 fields are stored for each row (month): date (*yyyy-MM-dd*), rainfall (mm), minimum temperature (ºC), and maximum temperature (ºC).  It is important to remind that data from all the weather stations under the selected database are merged before dumping them on the file following the average data fusion approach, which can be considered as a spatial fusion in this case. Note that date field values represent the day "01" of each month. The day is irrelevant since the important information is the month and the year. Note that "?" symbol is used in those cases where the value cannot be calculated due to missing values. This process is very fast (less than 1 second) and the size of the resulting file depends on the number of selected years (a few KiloBytes).

**Figure 9**. Minable View Creation functionality interface.

In the **"neighbour-based" analysis** the idea is a bit different. In this case, we intend to create a minable view, not with data from a particular whole database, but from a series of stations. The idea behind this analysis is that we may want to make a prediction for a certain station but the number of missing values is so high in that station that the analysis becomes nearly impossible. In those cases, we would like to train a model with data from other neighbor stations with better data quality in terms of lower missing value rate. The neighbor stations may belong to a different bordering country or even continent; we could even need to discard some particular neighbor stations if they also have the same missing values problem. That is why the selection of neighbor stations needs to be flexible and be done station by station. Since this type of analysis is more complex, we will only focus on a particular weather variable.

To create a minable view under this perspective, the user will select the range of years (From-To in Figure 9) and the weather variable (for temperature variables, we may choose between °C or °F) to consider for creating the minable view. Then, the user will select a particular station database (only station databases are allowed in this case as explained in the previous paragraph) and click on "Add neighbour station" button (Figure 9), repeating this step for each station that the user intends to include in the minable view. After incorporating all the desired stations, the user will select the option "Create .ARFF for standard analysis" and click on the button "Send", as depicted in Figure 9.

Then, the system will read from MongoDB the data from all the particular station databases (and the years and variable selected) and will create the .ARFF file. In this analysis, data from the selected stations are not fused spatially. On the contrary, monthly data for all the stations are stored in the resulting .ARFF file, so there will be one record for each month and neighbor station. As shown in Appendix I.B, the structure of the file is also different, storing, in this case, the following fields: year, month, value for the selected variable, the latitude of the station, longitude, and altitude (meters above sea level). These 3 last fields are obtained from the pre-loaded station database (see D. DATABASE CREATION). The size of the .ARFF file may be slightly higher in this type of analysis and will depend on the number of neighbor stations used (a few KiloBytes).

Note that the component in charge of implementing this functionality makes use of packages java.util, java.io, javax.swing, java.text, com.mongodb, org.bson and weka.core. In particular, this last package provides an API for integration with Weka methods, including data preparation and data analysis.

G. FORECASTING

This may be considered as the main module of the system, in which the prediction is performed. To do so, the user will select the kind of analysis ("Select kind of analysis" in Figure 10) and press the button "Choose". After that, depending on the type of analysis, the user will select the method and variable (if applicable) and will click on the button "Forecast". Then, the system will ask the user to indicate the minable view .ARFF file to be used and then, the predictions made will be shown in the table (result panel) on the right side (Figure 10). Note that the prediction will be made for the next 12 months to the period of time indicated when creating the minable view. For instance, if we created a minable view for years 2016, 2017, and 2018, the values shown on the table will correspond to the prediction made for the 12 months of the year 2019. Also, note that the result panel can be cleared at any time by pressing the button "Clear data".

As mentioned in the previous module, we can conduct a standard or a neighbour-based analysis. In turn, the standard analysis may be univariate or multivariate. Therefore, we may perform three different kinds of predictions that we will describe next. Regardless of the analysis, note that this module has been implemented in several classes that make use of different packages, among which it is important to highlight the Weka packages and methods that will be mentioned next.

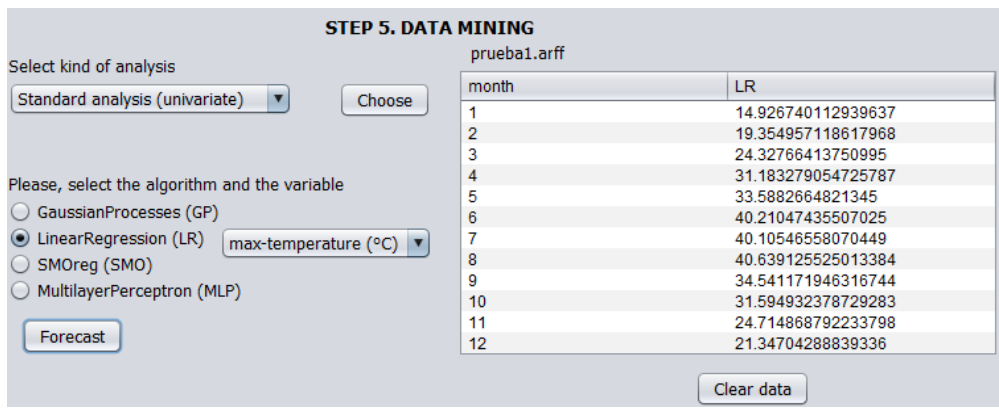

**Figure 10**. Forecasting functionality interface.

Regarding the **standard analysis**:

- If the user selects the option "Standard analysis (univariate)", he or she will be able to choose the predictive algorithm to be used: GP, LR, SMO, or MLP (see section 2); and the variable to predict. After that, the user has to indicate the minable view to be used, and then the prediction will be shown in the results panel. Note that in this kind of analysis the variable to forecast is predicted by using the minable view historical data from that particular variable only. To do so, we make use of the package

*weka.classifiers.timeseries.WekaForecaster*[††]. In particular, it is necessary to create an object from the class *WekaForecaster*, which provides a series of methods for performing the prediction. Then, we can set the variable to predict by using the method *setFieldsToForecast()*. Next, we will define the algorithm to be used by using the method *setBaseForecaster()*. After that, we invoke the method *forecast()* and the prediction is received on a List that is displayed on the results panel.

- If the user selects the option "Standard analysis (multivariate)" the process is quite similar with the only difference that, in this case, all the minable view historical data for all the variables will be used for the prediction of a particular variable. For instance, if we intend to predict the maximum temperature variable, not only historical data of maximum temperature variable will be used but also data from the rest of the variables (in this example, minimum temperature and rainfall) will be used to train the model. When implementing this approach, the only difference is that we will include all the variables by using the method *setFieldsToForecast()* and will filter the results for the variable we are interested in forecasting. The idea behind this analysis is that, due to data correlation, it may be possible to perform an accurate prediction of a variable using data from all the variables, which may be useful in cases of missing values for a particular variable. This approach is aligned with information fusion philosophy.

For **neighbour-based analysis**, also denoted as NBA, the process is quite different. In this case, after selecting the option (Neighbour-based analysis) and clicking on the "Choose" button, the user will select the particular weather station for which the prediction will be made (remind that this is a kind of analysis in which we have incorporated to the minable view data from different stations in order to make a prediction in a neighbour station, typically when the latter has a high rate of missing values). Then, he or she will click on the button "Forecast" and, after indicating the minable view .ARFF file to be used, the prediction will be shown in the results panel. Note that the user can easily select the prediction weather station by filtering station using the same menus (region, country, station) implemented for module "D. DATABASE CREATION" (see Figure 4). In this case, the user does not need to select the variable to predict since the minable view for this kind of analysis was created with data of a unique particular variable (see the previous module). For this type of analysis, we do not have to indicate the algorithm to be used. On the contrary, we have decided to use a Bagging (Bootstrap Aggregating) approach [47], which is a machine learning ensemble meta-algorithm with excellent results in terms of accuracy.

The Weka package used in NBA, called *weka.classifiers.meta.Bagging*[‡‡], works differently as well. Particularly, before running the Bagging method, the system needs to automatically create a test .ARFF file with data of the prediction station (see Appendix I.C). In that file, the system will include the rows for the months to predict and will use the symbol "?" to indicate which are the values to predict. The system also includes the latitude, longitude, and altitude for the prediction station. Note that this recently created .ARFF file will be the *test_set* while the minable view .ARFF file created in the previous module will be the *training_set*. Also, note that the .ARFF test file is only created for Weka package requirements and can be removed after analysis.

---

[††] https://weka.sourceforge.io/doc.packages/timeseriesForecasting/weka/classifiers/timeseries/WekaForecaster.html

[‡‡] https://weka.sourceforge.io/doc.dev/weka/classifiers/meta/Bagging.html

After that, the Weka package is ready to perform the prediction. We just need to create an object, denoted *cls*, from the class *Classifier* by invoking the method *new Bagging()*, and then train and test the model. The training of the model is performed by using the method *cls.buildClassifier(training_set)*, and the test is performing by creating a new object, called *eval*, from the class *Evaluation* and invoking the method *eval.evaluateModel(cls,test_set)*.

Note that in NBA the prediction is made by using not only weather data from neighbour stations but only geographical data such as latitude, altitude, and longitude. Those later variables are taken into account when training the model so the prediction can be more accurate. Again, this procedure can be considered as an information fusion approach.

## 4. System assessment

In this section, we will present some details about the tests and experiments conducted in order to validate our proposal. First, we will describe the usability tests conducted in order to improve the usability of the system (section 4.1). Then, we will present the details of a validation process that was carried out with the help of some experts in the area of weather prediction (section 4.2). Regarding ethics when dealing with experts/users and collecting data from them, we have followed some standard good practices consisting on providing them with comprehensive information about the project, asking them for permission, trying to avoid anything thay may cause them harm, and keeping confidentiality of information at every stage. We have also commited to inform them about the results once they are published. There was not any issue with respect to age or gender.

Finally, we will explain the performance results obtained by the system in terms of forecasting ability (section 4.3).

### 4.1. Usability tests

In this section, we will describe the process we followed in order to improve the usability of the system. In the first stage of the project, we implemented a preliminary version in which we did not formally consider usability issues. As a result, we obtained the prototype shown in Figure 11. As we can see, there is a single window and all the elements are mixed in a way that it is hard for the user to distinguish between the different functionalities of the system and the order in which each should be performed. We can also note that there are large empty areas of the main windows that are wasted and some overlapped buttons.

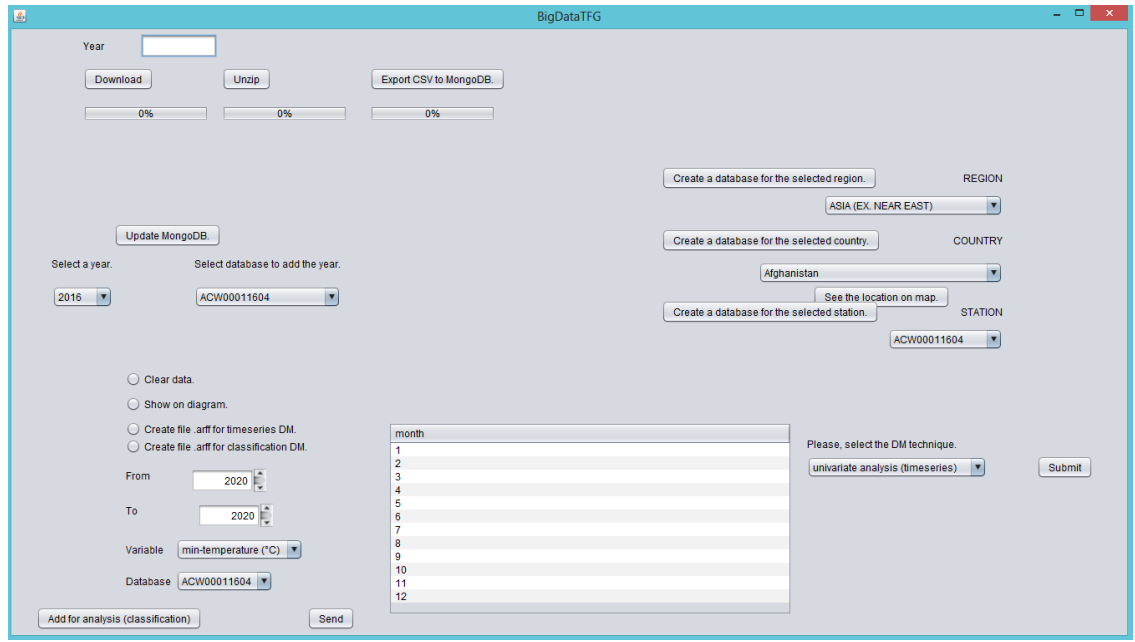

**Figure 11**. System preliminary interface.

After implementing that preliminary version, the system was evaluated by a group of users in terms of usability. In particular, we counted on 7 users aged between 21 and 34, with academic degrees of BSc (3 out of 7) and MSc (4 out of 7), and whose areas of knowledge are Computer Science (5 out of 7), Plant Production (1 out of 7) and Mathematics (1 out of 7). They were surveyed utilizing a series of questions shown in Table I, in which we should distinguish between items intended to measure the users' background and knowledge in the area (Items #1 to #12) and those intended to measure the users' assessment of the system (Items #13 to #15).

**Table I**. List of questions utilized for users' survey.

| Scope | #Item | Item Description | Answer range |
|---|---|---|---|
| Background | 1 | What is your knowledge level of technology in general? (1 = completely unfamiliar) | 1/2/3/4/5 |
| | 2 | How familiar are you with using computers? (1 = completely unfamiliar) | 1/2/3/4/5 |
| | 3 | How familiar are you with using computer applications? (1 = completely unfamiliar) | 1/2/3/4/5 |
| | 4 | How familiar are you with using Windows operating systems? (1 = completely unfamiliar) | 1/2/3/4/5 |
| | 5 | How familiar are you with downloading files? (1 = completely unfamiliar) | 1/2/3/4/5 |
| | 6 | How familiar are you with unzipping files? (1 = completely unfamiliar) | 1/2/3/4/5 |
| | 7 | How familiar are you with creating databases? (1 = completely unfamiliar) | 1/2/3/4/5 |
| | 8 | How familiar are you with Geography? (1 = completely unfamiliar) | 1/2/3/4/5 |
| | 9 | What is your knowledge about the study of rainfalls and temperature issues? (1 = completely unaware of) | 1/2/3/4/5 |
| | 10 | How familiar are you with times series analysis? (1 = completely unfamiliar) | 1/2/3/4/5 |
| | 11 | How familiar are you with the Data Mining area? (1 = completely unfamiliar) | 1/2/3/4/5 |
| | 12 | Have you ever used similar applications in the past? | Yes/No |
| Assessment | 13 | How far you do consider you managed to carry out the proposed tasks properly? (1 = completely unable to carry out) | 1/2/3/4/5 |
| | 14 | How useful did you find the system? (1 = completely useless) | 1/2/3/4/5 |
| | 15 | Would you use the system in a real environment if you had the chance? | Yes/No |

Regarding the background items, in this kind of process, it is crucial to provide information about the users, which may be important in order to contextualize the assessment and try to study its reproducibility. That is the reason why we decided to survey the users to collect information that could influence the results of the tests with respect to their familiarity to the main topics involved in this project (technology, computers, databases, window-oriented systems, Geography, files management, weather, data mining, and so on) and checked their expertise with similar systems in the past. After analyzing the results, we can conclude that most of the users have not used similar systems previously (85.7% versus 14.3%). Regarding the results obtained for numerical rated items (#1 to #11), those are shown in the box-plot shown in Figure 12. If we have a look at the items, we can notice a high level of expertise (Items #1 to #8) excluding weather-related and data mining issues (Items #9, #10, and #11).

Before taking the usability tests, the users were provided with a document with a general description of the system and some instructions on the assessment process. During the tests, the uses were asked to perform a series of tasks using the system with the supervision of a project member who took note of the main issues raised by the users. Those tasks are included and described in Table II, in which we can see that the main functionalities of the system have been widely tested by the users.

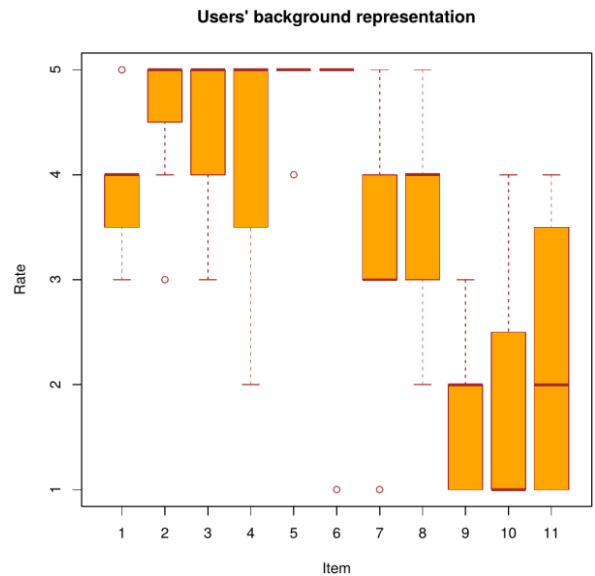

**Figure 12**. Users' background representation.

**Table II**. List of tasks performed by the users during usability tests.

| #Task | Task description |
|---|---|
| 1 | Download the global data related to the years 2016 to 2018 and save the corresponding files in D:\SystemTest\. Unzip the selected years' data and save them in D: \SystemTest\. Load the unzipped files into MongoDB. |
| 2 | Create a database for the particular station in Jordan located in "H4 Air Base" (Al Mafraq region), for which you can use the station map locator. Also, create a database for the particular station named "Shams Ma'an" located in Jordan and finally also create a database for Jordan as a country. |
| 3 | Load the data from 2016 to 2018 into the Jordan database recently created. Do the same with the "Shams Ma'an" database and with the H4 Air Base database. |
| 4 | Show a diagram for the minimum temperature in Celsius degrees from 2016 to 2018 for the Jordan database. Also, show a diagram for the rainfall from 2016 to 2018 for the "Shams Ma'an" database. |
| 5 | Create an ARRF file named "JordanRainfall.arff" for time series rainfall prediction purpose form 2016 to 2018 for the Jordan database and save it in the D:\systemTest folder. Also, do the same for "JOM00040310" (Shams Ma'an) station database with a file named "ShamsMaanMinC.arff" for time series minimum temperature in |
|   | Celsius prediction purpose. |
| 6 | By using the "JordanRainfall" file which created in task #5 and by using the *GaussianProcess* algorithm, show the Rainfall prediction for 2019 using only rainfall previous data. Do the same by using the three other algorithms: LinearRegression, SMOreg, and MultilayerPerceptron. |
| 7 | Clear the created displayed data generated from the previous step (Step # 6); then, by using the "ShamsMaanMinC" file which created in the task #5 and by using *GaussianProcess* algorithm, show the minimum temperature (in Celsius degrees) prediction for 2019 using the other variables (minimum temperature, maximum temperature, and rainfall) previous data. Do the same by using the other three algorithms: LinearRegression, SMOreg, and MultilayerPerceptron. |
| 8 | Repeat the previous task (task #7) 4 more times to predict each of the other four variables: minimum temperature in Fahrenheit, the maximum temperature in Celsius, the maximum temperature in Fahrenheit, and rainfall. |
| 9 | As you noticed in task #4, the percentage of complete data for rainfall for "JOM00040310" (Shams Ma'an) station database is very small. Therefore, Use the data from 2016 to 2018 of the nearby H4 Air Base database station (JOM00040250) in order to better predict the rainfall for 2019 in Shams Ma'an station. To do such a classification, you will have to create a new ".ARFF" file (named "H4AirBaseDataforShamsMaan.arff"), but before, you have to add the station H4 Air Base data for analysis. After that, you can create the mentioned .arff file and then, use it for prediction (classification) of rainfall in Shams Ma'an station. |

All of the usability tests took place at Jordan University of Science and Technology (Jordan, Irbid) on the period from 24th Nov. to 7th Dec. 2019. All the tests were performed using a laptop with a 15.6-inch display, Intel Core i5-5200u CPU, $1366 \times 768$ screen resolution, and Windows 7 operating system.

Regarding the assessment items of Table I, particularly focusing on Item #15, we could check that most users would use our system if they had the chance (85.7% versus 14.3%). Meanwhile, the users considered, on average, that they conducted the tasks properly (3.29/5) and found the system useful (3.14/5). Despite the number of users is not very high as to conduct more complex statistical proof, we can see that items reveal positive feedback from users for the usefulness of the system.

After conducting the tests and analyzing the results obtained, a series of aspects of improvement were detected and defined as shown in Table III. In that table, we have grouped the improvement aspects by functionalities of the system and also included a category with overall recommendations. We can see that most usability problems found by the users are due to an inconvenient format, location, or organization of the elements on the interface and the lack of some important information and messages.

**Table III**. Aspects of improvement raised by the users.

| Scope | Description |
|---|---|
| Overall | ▪ Add a title to the system's main interface<br>▪ Size up the used font size<br>▪ All the elements of the system must appear either if the user decides to "minimize" the window or "restore down"<br>▪ Group the sets of components by functionalities |
| Downloading panel | ▪ Add an instruction informing the user that he needs to download each year separately and therefore he cannot insert a range of years in the year textbox<br>▪ Add dialog messages, in addition to the status bars, informing the user that the task was performed successfully<br>▪ Add a sequential number indicating the order of the steps should be followed |
| Database creation panel | ▪ Swap the order of the "creating databases" buttons and filtering list combo-boxes<br>▪ Make the "see location on the map" button visible either if the user decides to "maximize" or "restore down" the system window<br>▪ Add a dialog message informing the user that the selected database was created successfully |
| Loading database panel | ▪ Move the "Update MongoDB" button to be below both "select a year" and "select database to add the year" list combo-boxes<br>▪ Add a status bar (like that used in Downloading Data panel) indicating that the loading data is in progress after the "Update MongoDB "button is clicked<br>▪ Show a dialog message after the completion of loading data informing the user that the database is loaded to MongoDB successfully |

| Minable view panel | ▪ Turn the radio button "clear data" into a button<br>▪ Replace all "DM" abbreviation to "Data Mining"<br>▪ Move the "Clear Data" button to be below the results panel<br>▪ Add dialog message informing the user that the particular .arff file was created successfully<br>▪ Add a dialog box informing that the particular station was added for analysis in the neighbor-based analysis |
|---|---|
| Data Mining panel | ▪ Swap the locations of results panel and data mining techniques panel<br>▪ Translate all the error messages that might be appeared into English (when the startup language was English)<br>▪ When the results are shown in the result panel, add the name of the selected minable view file |

Finally, the second round of implementation was carried out in order to address all the issues detected after usability tests. As a result, a new more stable, usable, and robust version of the system was obtained, as shown in Figure 13. In this case, we can notice that the wasted space at the interface has been reduced and the organization of the element is much cleared. In addition, the main groups of functionalities have been separated and ordered. This way, the process of use of the system by the users becomes easier as these improvements are based on the recommendations made by the users themselves.

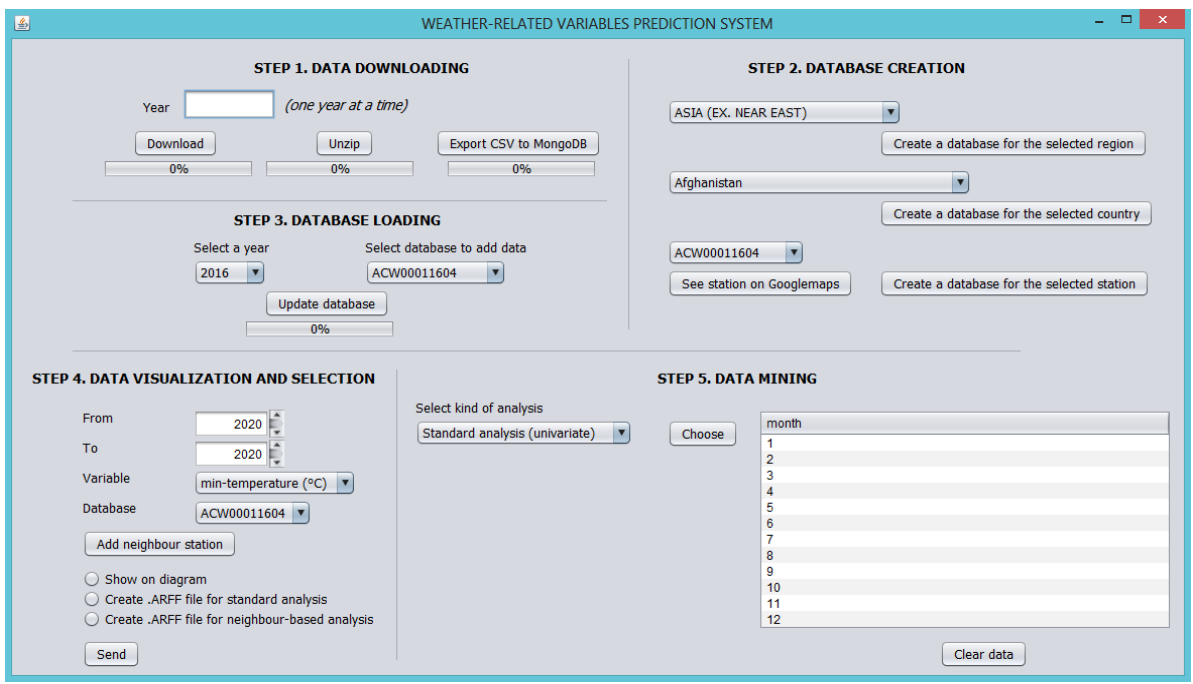

**Figure 13**. System interface after improvements.

### 4.2. Expert validation

After conducting the usability tests, some experts on the domain of application were surveyed about their background (Item #1 and #2 in Table IV) and opinion (Items #3 to #9 in Table IV) with respect to the system. In total, we relied on 7 experts aged between 42 and 58, all of whom are experts for 18 years (average) in Natural Resources, teaching and researching in that area as professors in university high education.

**Table IV**. List of questions utilized for experts' survey.

| Scope | #Item | Item Description | Answer range |
|---|---|---|---|
| Background | 1 | What is your knowledge about the study of rainfalls and temperature issues (1 = completely unfamiliar)? | 1/2/3/4/5 |
| | 2 | Have you ever used similar applications in the past? | Yes/No |
| Assessment | 3 | How far you do consider that the shown tasks were carried out properly in the system? (1 = completely improperly)? | 1/2/3/4/5 |
| | 4 | How useful did you find the system? (1 = completely useless)? | 1/2/3/4/5 |
| | 5 | In general, how would you assess the system in terms of ease of use and understanding? (1 = the worst value)? | 1/2/3/4/5 |
| | 6 | How far did you find that the graphic design of the system is helpful to accomplish the specified tasks? (1 = completely inconvenient)? | 1/2/3/4/5 |
| | 7 | How far did you find that the various functionalities in the system are properly integrated? (1 = completely inappropriately)? | 1/2/3/4/5 |
| | 8 | Did the system meet your expectations about it? (1 = completely not)? | 1/2/3/4/5 |
| | 9 | Would you recommend using the system in a real environment if you had the chance? (1 = not recommend at all) | 1/2/3/4/5 |

Regarding the background items, again it is important to collect this kind of information in order to contextualize the results and provide important information that could let other authors conduct equivalent analysis. In particular, the experts were questioned about their degree of knowledge on the domain of reference and their expertise with using similar systems in the past. We obtained an average value of 4.14 in Item #1, which demonstrate a real knowledge of experts in the area of application of our system. In addition, 3 out of the 7 experts had used similar applications in the past.

In this experiment, we showed the users how the system works and performed in front of them the same tasks considered in the usability tests. This process took place at Jordan University of Science and Technology (Jordan, Irbid) on the period from 10th Dec. to 30th Dec. 2019. All the tests were performed using a laptop with a 15.6-inch display, Intel Core i5-5200u CPU, 1366 × 768 screen resolution, and Windows 7 operating system. Before the assessment, a document was provided to the experts in advance so that they could have some overall information about the system and the process to be conducted.

After that, the experts were surveyed about their opinion on the system by using assessment items #3 to #9 from Table IV. The results obtained are presented in the box-plot displayed in Figure 14. Despite the number of experts is not very high as to conduct a more advances statistical proof, we can see that experts have rated our system positively in all the items (particularly concerning the ease of use), except perhaps in item #6 which is related to the graphic design of the system that seems to be an aspect to be improved.

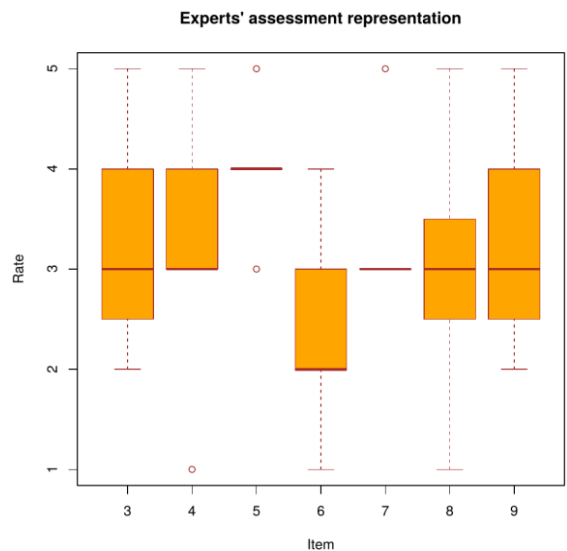

**Figure 14**. Experts' assessment representation.

Finally, note that the experts were also queried about the system's weak/strong points and potential future lines of development and research. The results obtained will be commented in Section 5 for the sake of document organization coherence.

**4.3. Predictive performance**

In this section, we will explain the experiments that we have conducted to assess the predictive performance of our system. It is important to clarify that the purpose of this section is to present some scenarios or test cases that demonstrate the applicability and usefulness of our approach.

*4.3.1. Evaluation metrics*

In order to evaluate the predictive performance of our system, we decided to use two metrics: (1) The Normalized Mean Squared Error (NMSE), which measures the differences between the actual and the predicted values, as shown in Eq. (1); (2) the Directional Symmetry (DS), which represents the performance when predicting the direction of change, positive or negative, of a time series, as shown in Eq (2).

$$NMSE = \frac{1}{n} * \frac{\sum_{i=1}^{n}(x_i - x'_i)^2}{\bar{x} * \bar{x}'} \quad (1)$$

$$DS = \frac{1}{n-1} * \sum_{i=2}^{n} d_i, \text{ where } d_i = \begin{cases} 1, if (x_i - x_{i-1}) * (x'_i - x'_{i-1}) \geq 0 \\ 0, otherwise \end{cases} \quad (2)$$

If both formulas, *X* represents the actual variable to predict, *X'* represents the predicted value variable and *n* is the number of observations.

We have decided to use these two metrics as they have already been used with success in similar previous projects [44] and that would let us carry out some kind of comparison with those previous works.

*4.3.2. Case definition*

We have designed three groups of experiments to test: A) the performance of standard univariate analysis; B) the performance of the multivariate analysis compared to univariate analysis; and, C) the performance of neigbour-based analysis approach compared to univariate and multivariate approaches.

Table V contains an explanation of all the tests, in which we have varied different parameters: group of experiments, training set period, test set period, variable to predict, geographical area, and techniques.

**Table V**. Test case definitions.

| Group | No. | Variable | Training period | Test period | Geographical area | Techniques |
|---|---|---|---|---|---|---|
| A | 1 | Precipitation | 1933-2018 | Jan-Dec 2019 | Castilla y Leon (Spain) | GP, LR, SMO |
| A | 2 | Max. Temperature | 2016-2018 | Jan-Dec 2019 | Córdoba (Spain) | GP, LR, SMO |
| B | 3 | Precipitation | 2010-2017 | Jan-Jun 2018 | Cisjordania | GP |
| B | 4 | Min. Temperature | 2010-2017 | Jan-Jun 2018 | Jordan | GP |
| C | 5 | Precipitation | 2010-2017 | Jan-Jun 2018 | Mafraq (Jordan) | GP |
| C | 6 | Min. Temperature | 2010-2017 | Jan-Jun 2018 | Mafraq (Jordan) | GP |

Note that we have not used the MLP for the experiments since the highly preliminary inaccurate results obtained led us not to consider this technique appropriate for this purpose. In cases 3 to 6, in which a complex comparison between approaches needs to be conducted, we decided to only experiment with GP technique for it being the one with the best results in case 1. In cases 5 and 6, the purpose is to predict in King Hussein Air Base (Mafraq, Jordan) station area, by using data from the four remaining stations of Jordan. Also note that we have tried to vary the scenarios as far as possible, in order to cover different geographical areas, lengths periods of time and variables. This way we can have a wider impression of the system.

*4.3.3. Results*

With respect to group A, test number 1 was performed and the results obtained are depicted in Figure 15, which includes the predicted values by techniques GP, LR and SMO, and the actual values. It is important to highlight that the methods seem to predict the variations of the series although some deviations are noticeable. In particular, we can see that the actual spring precipitations peak seems to have occurred around 1 month in advance with respect to the predictions. Also, note that the 2019 summer was atypically rainy in the analyzed area. Finally, the system predicts the actual autumn precipitations peak with some deviations in terms of amount.

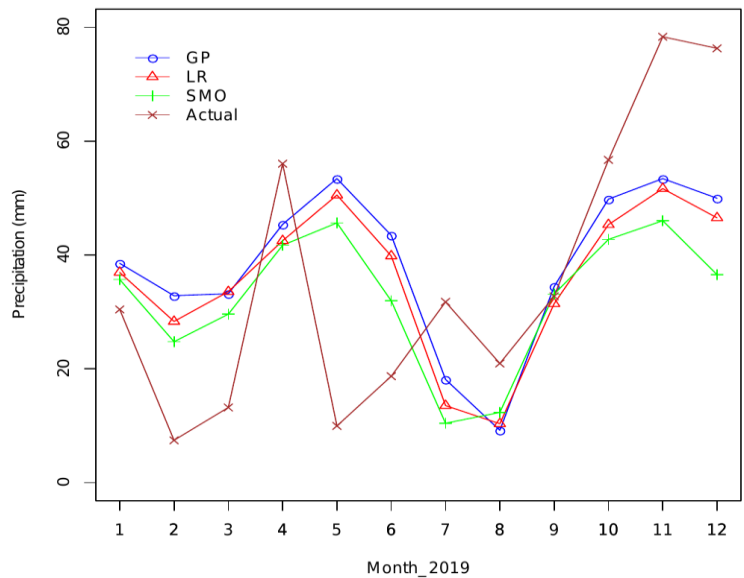

**Figure 15**. Actual versus predicted values in test case 1.

In order to measure the performance of the system quantitatively, we have calculated the NMSE and DS indicators, and the results are presented in Table VI (Month 1= January, 2=February, and so on). The best results in each case are represented as bold. Also, note that for the DS indicator we have only represented the variation (+ or -) and the final result. Analyzing the results, we notice that GP is the technique that obtains the best overall results in terms of NSME. In addition, GP is the best technique in 5 out of the 12 months, while SMO is the best in 5 out of the 6 months, and LR only is considered as the most accurate technique in 1 out of the 12 months. Regarding DS indication, both GP and LR obtain a remarkable value of 0.727, which indicates that the variations of the series and quite well predicted. DS indicator value is a bit lower for SMO, 0.636.

**Table VI**. Results for case 1.

| Month | NMSE | | | DS | | |
|---|---|---|---|---|---|---|
| | GP | LR | SMO | GP | LR | SMO |
| 1 | 0.00394999 | 0.00272998 | **0.00196431** | | | |
| 2 | 0.03876261 | 0.02809951 | **0.02131248** | + | + | + |
| 3 | 0.02404428 | 0.02670433 | **0.01912011** | + | + | + |
| 4 | **0.00681202** | 0.01179568 | 0.01456589 | + | + | + |
| 5 | 0.11350094 | 0.10610484 | **0.09046734** | - | - | - |
| 6 | 0.03685188 | **0.0287931** | 0.01248378 | - | - | - |
| 7 | **0.01115258** | 0.02140747 | 0.03230968 | - | - | - |
| 8 | 0.00829388 | 0.00724077 | **0.00526578** | + | + | - |
| 9 | 0.00017664 | 0.00010986 | **0.00001051** | + | + | + |
| 10 | **0.00285331** | 0.00833471 | 0.0138696 | + | + | + |
| 11 | **0.0373051** | 0.04587286 | 0.07420982 | + | + | + |
| 12 | **0.04169255** | 0.05720953 | 0.11239479 | + | + | + |
| Overall | **0.02711631** | 0.02870022 | 0.03316451 | **0.727** | **0.727** | 0.636 |
| | Avg | | | Result | | |

With respect to case 2 (prediction of maximum temperature at Cordoba Airport - Spain), the results obtained are shown in Figure 16, which includes the actual and the predicted values by techniques GP, LR and SMO. In this case we can see that the results obtained by LR are far from positive since the regression function is a constant line that does not fit the curve properly due to the intrinsic limitations of linear regression functions when dealing with periodical time series. However, the results obtained by GP and SMO are really accurate since both techniques seem to fit the actual values line during the whole year.

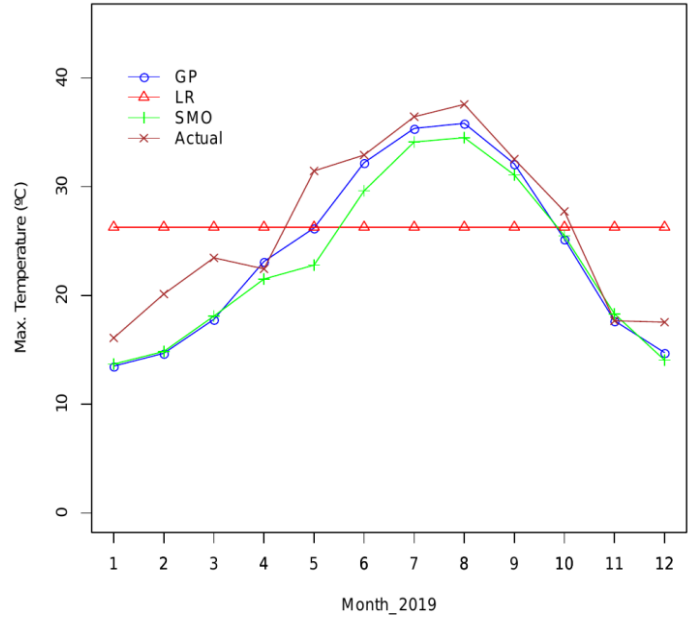

**Figure 16**. Actual versus predicted values in test case 2.

Table VIII contains the quantitative results of case 2 for the metrics considered. Regarding NMSE, we can see really good results for GP technique, which is the best in 10 out of the 12 months of the year, and it is also the best overall obtaining a result of around 0.00058. With respect to DS indicator, both GP and SMO obtain very good values (0.909 in both cases), which demonstrates that the predicted curves reflect the trend of the actual series almost perfectly. In this case, the DS indicator does not make any sense for LR due to the constant nature of regression line.

**Table VII**. Results for case 2.

|  | NMSE |  |  | DS |  |  |
| :---: | :---: | :---: | :---: | :---: | :---: | :---: |
| Month | GP | LR | SMO | GP | LR | SMO |
| 1 | **0.00040371** | 0.00669346 | 0.00040971 |  |  |  |
| 2 | **0.00178357** | 0.00244207 | 0.00198124 | + | NA | + |
| 3 | 0.00192429 | **0.00051445** | 0.00203749 | + | NA | + |
| 4 | **2.7816E-05** | 0.00095053 | 6.1821E-05 | - | NA | - |
| 5 | **0.00165662** | 0.00172299 | 0.00532438 | + | NA | + |
| 6 | **2.9078E-05** | 0.00283354 | 0.00076887 | + | NA | + |
| 7 | **6.8988E-05** | 0.00664102 | 0.00038186 | + | NA | + |
| 8 | **0.00018112** | 0.00821657 | 0.00066696 | + | NA | + |
| 9 | **1.1472E-05** | 0.00251804 | 0.0001489 | + | NA | + |
| 10 | 0.00038848 | **0.00013553** | 0.00035648 | + | NA | + |
| 11 | **2.1142E-09** | 0.00477869 | 2.8641E-05 | + | NA | + |
| 12 | **0.00047241** | 0.00492408 | 0.00086603 | + | NA | + |
| Overall | **0.00057896** | 0.00353091 | 0.00108603 | **0.909** | NA | 0.909 |
|  |  | Avg |  |  | Result |  |

Regarding group B, we have conducted test cases 3 and 4. With respect to case 3, Figure 17 shows the predicted values for univariate and multivariate approaches and the actual values. We can see that both predictions adjust well to the trend of the series. However, there are some differences in the first two months of 2018, which were atypically rainy in Cisjordania. Anyway, it is noticeable that the multivariate approach fits more accurately with the actual series.

Table VIII contains information about the predictive power of the methods in this test case. In particular, it is noticeable that the multivariate approach improves the univariate approach in all months and globally in terms of NSME. The DS has a remarkable value in both cases, 0.8. This experiment shows that the multivariate analysis can be useful in certain scenarios, particularly in this scenario with a medium rate of missing values for precipitation variable (49.7% in this case).

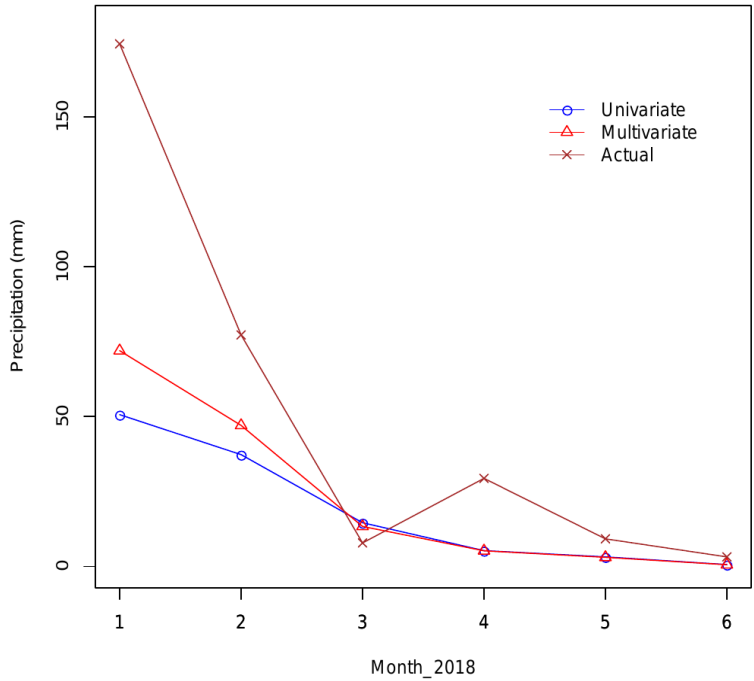

**Figure 17**. Actual versus predicted values in test case 3.

**Table VIII**. Results for case 3.

|  | NMSE | | DS | |
| --- | --- | --- | --- | --- |
| Month | Univariate | Multivariate | Univariate | Multivariate |
| 1 | 2.73981101 | **1.48165970** | | |
| 2 | 0.28554240 | **0.12882282** | + | + |
| 3 | 0.00814755 | **0.00428827** | + | + |
| 4 | 0.10412003 | **0.08258316** | - | - |
| 5 | 0.00648056 | **0.00537711** | + | + |
| 6 | 0.00118110 | **0.00096219** | + | + |
| Overall | 0.52421378 | **0.28394888** | 0.80 | 0.80 |
| | Avg | | Result | |

Concerning case 4, Figure 18 depicts the predicted and the actual values for minimum temperature. In this case, we can see that both predictions adjust quite well to the trend of the series and to the actual values (just a few degrees lower on average). It is noticeable that the multivariate approach fits more accurately with the actual series in the first part of the year (winter) and univariate obtains better results in the second part (spring).

If we have a look at the quantitative results, shown in Table IX, the mentioned above can be checked. Besides, we notice that multivariate analysis prediction is lightly more accurate overall in terms of NMSE (0.00038954 versus 0.00043352). Regarding DS indication, the multivariate approach perfectly reflects the trend of the series (1.00).

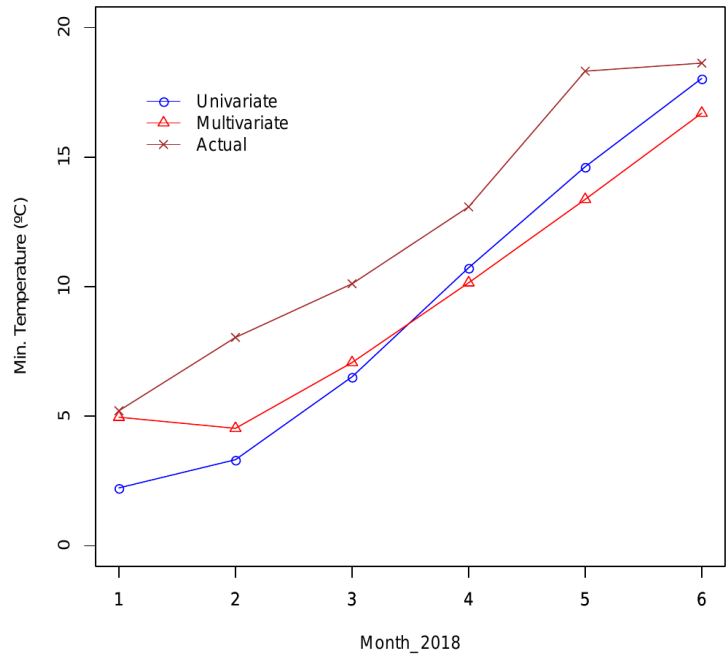

**Figure 18**. Actual versus predicted values in test case 4.

**Table IX**. Results for case 4.

|  | NMSE | | DS | |
|---|---|---|---|---|
| Month | Univariate | Multivariate | Univariate | Multivariate |
| 1 | 0.00036357 | **0.00000250** | | |
| 2 | 0.00091209 | **0.00049275** | + | - |
| 3 | 0.00052765 | **0.00036963** | + | + |
| 4 | **0.00022609** | 0.00034336 | + | + |
| 5 | **0.00055745** | 0.00098000 | + | + |
| 6 | **0.00001425** | 0.00014898 | + | + |
| Overall | 0.00043352 | **0.00038954** | **1.00** | 0.80 |
|  | Avg | | Result | |

Finally, regarding group C, two tests have been carried out. Regarding case 5, figure 19 shows the predicted values by using the different approaches (including the neighbour-based one, NBA) and the actual values. In this case, we can see that the NBA approach line is really close to the actual values line, particularly in the first part of the year. The univariate and multivariate approaches do not manage to fit well to the actual line in this case.

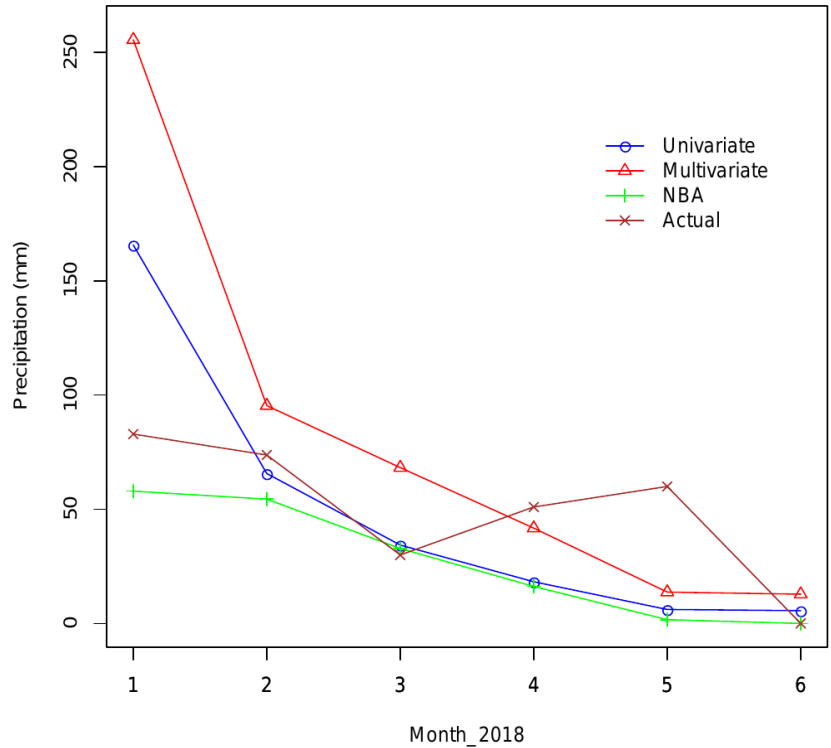

**Figure 19**. Actual versus predicted values in test case 5.

In search of a better understanding of this phenomenon, we can have a look at the quantitative results included in Table X. In this case, we see that NBA approach is the one that best fits the actual values in 3 out of the 6 considered months and also overall (0.11577066). The DS indicator shows that the trend of the series is not quite well represented by the models in this test case (0.6 for all the approaches). The results of this test, however, are quite positive since they prove that in scenarios like this, with a really high rate of missing values in precipitation variable (92.49%), the NBA approach designed in this work can be useful to achieve a more accurate prediction.

**Table X**. Results for case 5.

|  | NMSE |  |  | DS |  |  |
| :---: | :---: | :---: | :---: | :---: | :---: | :---: |
| Month | Univariate | Multivariate | NBA | Univariate | Multivariate | NBA |
| 1 | 0.46694595 | 1.23237293 | **0.07725206** |  |  |  |
| 2 | **0.00446085** | 0.01936530 | 0.04608872 | + | + | + |
| 3 | 0.00127790 | 0.06047075 | **0.00098993** | + | + | + |
| 4 | 0.07301511 | **0.00354594** | 0.14960253 | - | - | - |
| 5 | 0.19779470 | **0.08849543** | 0.42069071 | - | - | - |
| 6 | 0.00215275 | 0.00679651 | **0.00000000** | + | + | + |
| Overall | 0.12427454 | 0.23517448 | **0.11577066** | 0.60 | 0.60 | 0.60 |
|  |  | Avg |  |  | Result |  |

Regarding case 6, if we look at figure 20, it is noticeable that all approaches fit quite well the actual curve. During the first 3 months of the year, the multivariate approach seems to be the most accurate while univariate technique is the best during the second trimester. Curiously, in April (month 4) all approaches seem to perfectly predict the actual value.

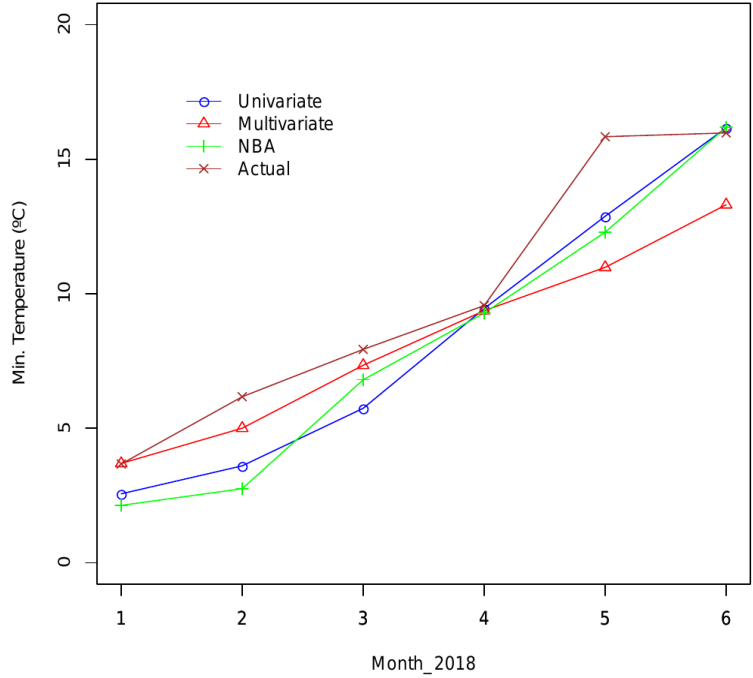

**Figure 20**. Actual versus predicted values in test case 6.

If we have a look at the quantitative results of case 6 in Table XI, we can confirm the above. In addition, we can check that multivariate approach is the most accurate overall. In this case, NBA approach does not improve the more naïve approaches at all. Regarding DS indicator, all the approaches perfectly represents the trend of the series (1.00 in all the cases).

**Table XI**. Results for case 6.

|         | NMSE       |              |            | DS         |              |      |
|---------|------------|--------------|------------|------------|--------------|------|
| Month   | Univariate | Multivariate | NBA        | Univariate | Multivariate | NBA  |
| 1       | 0.00008398 | **0.00000002** | 0.00029314 |            |              |      |
| 2       | 0.00045018 | **0.00005661** | 0.00144571 | +          | +            | +    |
| 3       | 0.00032690 | **0.00001440** | 0.00015783 | +          | +            | +    |
| 4       | **0.00000082** | 0.00000149 | 0.00000969 | +          | +            | +    |
| 5       | **0.00059718** | 0.00097675 | 0.00155771 | +          | +            | +    |
| 6       | **0.00000246** | 0.00029480 | 0.00000545 | +          | +            | +    |
| Overall | 0.00024359 | **0.00022401** | 0.00057826 | **1.00**   | **1.00**     | **1.00** |
|         | Avg        |              |            | Result     |              |      |

# 5. Discussion and applicability of the proposal

In the previous section we have shown the results obtained when assessing our system in a series of scenarios with the purpose of presenting a few interesting situations and check how the system performs. A summary of the best results obtained in each case can be found in Table XII.

We do not intend to comprehensively compare our approach with other existing approaches because, as stated in Section 2, many of the related works found in the literature do not include experimental results but only focus on computational cost. The ones that include a quantitative validation cannot be completely directly to our approach since they use different variables and granularity levels.

Nevertheless, we will calculate some indicators to compare our approach with the work presented by Lin et al. in [44], which is the most similar work that includes a quantitative assessment. From the table XII, we can conclude that our approach outperforms the method by Lin et al. in cases 2, 4, 6 and overall in terms of NMSE, and it has a more accurate behavior in cases 2, 3, 4, 6 and overall in terms of DS. In fact, the case 4 would be the most appropriate for which a certain direct comparison can be conducted with the work presented by Lin et al., where authors predict the same variable as our method (temperature in this case). However, the main difference with our approach is that the authors predict with an hourly granularity while we do it on a monthly basis. It may seem that working on an hourly granularity is harder for prediction purposes, and that may be true, but it is also true that the training set is also more precise so there may be offset. Anyway, as we mentioned in the previous paragraph, our method outperforms the one by Lin et al. in case 4 for both NMSE and DS metrics.

**Table XII**. Summary of (best) results.

| Indicator | Case 1 | Case 2 | Case 3 | Case 4 | Case 5 | Case 6 | Average | Lin et al. |
|---|---|---|---|---|---|---|---|---|
| NMSE | 0.0271163 | 0,0005789 | 0.2839488 | 0.0003895 | 0.1157706 | 0.0002240 | 0.0001338 | 0.0010424 |
| DS | 0.727 | 0.909 | 0.80 | 1.00 | 0.60 | 1.00 | 0.8393333 | 0.750551 |

If we have a look at table XII, we can extract different findings. First, it is not possible to conclude that any of the 3 kinds of analysis proposed in this paper is better than the rest because all of them present excellent results (case 1 in group A, case 4 in group B and case 6 in group C). We have also noticed that using longer periods of time for training helps to obtain better results (see case 1 in which we used training data from 85 years versus case cases 3 and 5 in which we used 8 years).

In our experiments we included predictions for 6 and 12 months, and in both scenarios we have obtained a variety of good and poorer results. In particular, it is important to mention the unexpected wrong behavior of linear regression in 12 months prediction in cases where a periodical behavior can be noticed in the time series. Regarding the geographical area, we have also obtained different types of results in Spain and Jordan, having good and poorer results in both cases. With respect to the technique to use, the results concluded that Gaussian Processes provide with more accurate results in the considered scenarios than Linear Regression, SVM or Neural Networks.

It is noticeable, however, that our system is more accurate when predicting maximum and minimum temperature than when predicting precipitation. This is definitely due to the fact that average (minimum and maximum) temperatures are more stable variables from one year to another but precipitation may considerably vary on a temporal margin of weeks from one year to another. That is undoubtedly the main limitation of our method that can be solved by considering a more specific granularity (weekly or daily) when forecasting precipitation.

In addition, our system limited to an implementation where we considered 3 weather variables with a monthly granularity: maximum temperature, minimum temperature, and rainfall. It takes data for a particular open data repository so the current implementation is dependable on the data quality exhibited by data in that repository. Our system has been run on a single computer as a prototype and there still much to do in order to take advantage of big data infrastructure and its computational power. However, the ideas of this system have presented in this paper in a way general enough so that other researchers can take advantage of them and reproduce or adapt for their particular implementations. Those ideas refer to the way data are captured, internally managed, and analyzed by a series of predictive techniques.

As strong points, we could highlight that the system performs predictive tasks that have not been observed in similar big data systems, i.e., the exploitation of correlation between variables and the neighborhood of stations for achieving more accurate results. Another strong point is the comprehensive validation that the system has been exposed to. We have validated from three different perspectives: usability aspects, experts' validation, and predictive performance. In all those three groups of experiments, very satisfactory results have been obtained that prove the validity and usefulness of our system.

## 6. Conclusions and future lines

In this paper, we have proposed a visual big data system for forecasting the values of some weather-related variables using information from weather stations all over the world. Our system takes advantage of open weather data repositories and imports it to a local MongoDB database where data are organized in collections so they are easier to manage. By using the multimedia resources of the system, the user can create flexible and customized minable views for later analysis, in which different data mining techniques can be used for predictive purposes. In particular, we have developed a solution that lets the user predict the values of a certain variable by using historical values of other related variables. Moreover, our system also can predict values for a certain region in which too few data are collected at the corresponding station by using the information of the neighbor stations with more complete data. These features are particularly useful in scenarios with a high rate of missing values.

Our system has been validated by a group of experts who found it valid in terms of usefulness and ease of use, among others. We have also conducted experiments in order to improve the usability of the system after implemented the detected aspects of improvement. Finally, we performed a quantitative assessment of the system in terms of predictive power, obtaining very satisfactory preliminary results that encourage us to keep working on an evolution of the system.

Our system limited to a prototypical implementation ran on a single computer and therefore a more realistic implementation needs to be developed. However, the ideas presented in this paper may be useful for facing similar projects, particularly those regarding the exploitation of the correlation between meteorological variables and the neighborhood of stations.

There is still a long way to cover in this project and many aspects and future lines could be addressed in the future. Some of the most promising are:

- To make the system more flexible so that it can read data from other repositories all over the world. This way, the chances of having missing values for a certain region would be reduced. A data fusion strategy could be useful for this purpose.

- Related to the above, it would be interesting to consider other weather-related variables, not only temperature or rainfall, but humidity, wind, and so on.

- It is very convenient to improve the system so that it can work with finer granularity. In its current version, a monthly granularity is implemented. Having a weekly, daily, or even finer granularity would be very useful for prediction, particularly in extreme weather conditions that tend to appear suddenly.

- Not only considering time series data but other global physical parameters (e.g. climate change, global warming, climate shifting, high spatial and temporal data variability, etc.) could be useful for prediction purposes.

- Regarding the assessment of the system, although very promising results have been obtained, it would be convenient to conduct similar experiments with a higher number of experts and users to statistically confirm the preliminary results. It is also important to conduct computational performance tests and carry out a series of experiments to prepare a big data infrastructure composed of a cluster of computers, for more global, parallel and efficient use of the prototype of the system shown in this paper.

**Acknowledgment:**

This paper was drafted as part of Juan A. Lara's research stay during 2019-2020 at Jordan University of Science and Technology, JUST (Jordan), which partially sponsored this research. The authors would like to thank UDIMA's and JUST's students who took part in the design and implementation of the system, particularly Francisco Javier Moreno Hermosilla, Paulina Pyzel and Amnah Al-Abdi; and JUST's experts for providing their feedback in order to assess this system.

**References**

[1] C. Lynch, Big data: How do your data grow? *Nature*, vol. 455, no. 7209, pp. 28–29, 2008.

[2] G. Firican, The 10 Vs of Big Data. TDWI. https://tdwi.org/articles/2017/02/08/10-vs-of-big-data.aspx [accessed July 2020]

[3] C. Küçükkeçeci and A. Yazici, Multilevel Object Tracking in Wireless Multimedia Sensor Networks for Surveillance Applications Using Graph-Based Big Data, *IEEE Access*, vol. 7, pp. 67818-67832, 2019.

[4] P. Kulkarni, and K. B. Akhilesh K.B, Big Data Analytics as an Enabler in Smart Governance for the Future Smart Cities. In: Akhilesh K., Möller D. (eds) Smart Technologies. Springer, Singapore, 2020.

[5] Y. Wu, H. Huang, N. Wu, Y. Wang, M. Z. A. Bhuiyan, and T. Wang, An incentive-based protection and recovery strategy for secure big data in social networks, *Information Sciences*, vol. 508, pp. 79-91, 2020.

[6] M. Ambigavathi, and D. Sridharan D, A Survey on Big Data in Healthcare Applications. In: Choudhury S., Mishra R., Mishra R., Kumar A. (eds) Intelligent Communication, Control and Devices. Advances in Intelligent Systems and Computing, vol 989. Springer, Singapore, 2020.

[7] K. Narendra, and G. Aghila, Securing Online Bank's Big Data Through Block Chain Technology: Cross-Border Transactions Security and Tracking. In R. Joshi, & B. Gupta (Eds.), Security, Privacy, and Forensics Issues in Big Data pp. 247-263, 2020.

[8] Fuad Bajaber, Sherif Sakr, Omar Batarfi, Abdulrahman Altalhi, Ahmed Barnawi, Benchmarking big data systems: A survey, *Computer Communications*, vol. 149, pp. 241-251, 2020.

[9] U. M. Fayyad, G. Piatetsky-Shapiro, and P. Smyth, "From Data Mining To Knowledge Discovery: An Overview," in *Advances In Knowledge Discovery And Data Mining*, eds. U.M. Fayyad, G. Piatetsky-Shapiro, P. Smyth, and R. Uthurusamy, AAAI Press/The MIT Press, Menlo Park, CA., pp. 1-34, 1996.

[10] A. Shastri, and M.Deshpande, A Review of Big Data and Its Applications in Healthcare and Public Sector. In: Kulkarni A. et al. (eds) Big Data Analytics in Healthcare. Studies in Big Data, vol. 66. Springer, Cham, 2020.

[11] A. Baerg, Big Data, Sport, and the Digital Divide: Theorizing How Athletes Might Respond to Big Data Monitoring. *Journal of Sport and Social Issues*, vol. 41, no. 1, pp. 3–20, 2017.

[12] R. Yang, L. Yu, Y. Zhao, H. Yu, G. Xu, Y. Wu, and Z. Liu, Big data analytics for financial Market volatility forecast based on support vector machine, *International Journal of Information Management*, vol. 50, pp. 452-462, 2020.

[13] E. Hussein, R. Sadiki, Y. Jafta, M. M. Sungay, O. Ajayi, and A. Bagula A., Big Data Processing Using Hadoop and Spark: The Case of Meteorology Data. In: Zitouni R., Agueh M., Houngue P., Soude H. (eds) e-Infrastructure and e-Services for Developing Countries. AFRICOMM 2019. Lecture Notes of the Institute for Computer Sciences, Social Informatics and Telecommunications Engineering, vol. 311. Springer, Cham, 2020.

[14] J. F. Torres, A. Troncoso, I. Koprinska, Z. Wang, and F. Martínez-Álvarez, Big data solar power forecasting based on deep learning and multiple data sources, *Expert Systems*. 36:e12394, 2019. https://doi.org/10.1111/exsy.12394.

[15] A. Corbellini, C. Mateos, A. Zunino, D. Godoy, and S. Schiaffino, Persisting big-data: The NoSQL landscape, *Information Systems*, vol. 63, pp. 1-23, 2017.

[16] F. Marchioni, Infinispan Data Grid Platform, Packt Pub Limited, Birmingham, UK, 2012.

[17] A. Lakshman, and P. Malik, Cassandra: a decentralized structured storage system, *ACM SIGOPS Oper. Syst. Rev.*, vol. 44, no. 2, pp. 35-40, 2010.

[18] Objectivity Inc., InfiniteGraph, http://www.objectivity.com/infinitegraph, 2013 (accessed 17.04.20).

[19] K. Chodorow, and M. Dirolf, MongoDB: The Definitive Guide, O′Reilly Media, Inc., Sebastopol, CA, USA, 2010.

[20] P. Membrey, E. Plugge, and T. Hawkins,The Definitive Guide to MongoDB: The NoSQL Database for Cloud and Desktop Computing, Apress, Berkely, CA, USA, 2010.

[21] B. Jose and S. Abraham, Exploring the merits of nosql: A study based on mongodb, *2017 International Conference on Networks & Advances in Computational Technologies (NetACT)*, Thiruvanthapuram, pp. 266-271, 2017.

[22] H. Hassani, and E. S. Silva, Forecasting with Big Data: A Review, *Ann. Data. Sci.*, vol.. 2, pp. 5–19, 2015.

[23] C. Aggarwal, Data Classification – Algorithms and Applications, Chapman & Hall/CRC, 2014.

[24] P. A. Gutiérrez, M. Pérez-Ortiz, J. Sánchez-Monedero, F. Fernández-Navarro and C. Hervás-Martínez, Ordinal Regression Methods: Survey and Experimental Study, in *IEEE Transactions on Knowledge and Data Engineering*, vol. 28, no. 1, pp. 127-146, 2016.

[25] H. Liu, Y. Ong, X. Shen and J. Cai, When Gaussian Process Meets Big Data: A Review of Scalable GPs, in *IEEE Transactions on Neural Networks and Learning Systems*.

[26] S. Haykin, Neural Networks: A Comprehensive Foundation (2 ed.). Prentice Hall, 1998.

[27] S. K. Shevade, S. S. Keerthi, C. Bhattacharyya, and K. R. K. Murthy, Improvements to the SMO Algorithm for SVM Regression, *IEEE Transactions on Neural Networks*, 1999.

[28] G. A. F. Seber, and A. J. Lee, Linear Regression Analysis, $2^{nd}$ edition, Wiley Series in Probability and Statistics, Wiley-Interscience, 2003.

[29] P. Pandey, M. Kumar and P. Srivastava, Classification techniques for big data: A survey, *2016 3rd International Conference on Computing for Sustainable Global Development (INDIACom)*, New Delhi, pp. 3625-3629, 2016.

[30] D. Renuka Devi, and S. Sasikala, Online Feature Selection (OFS) with Accelerated Bat Algorithm (ABA) and Ensemble Incremental Deep Multiple Layer Perceptron (EIDMLP) for big data streams. *Journal of Big Data,* vol. 6, no. 103, 2019.

[31] I. H. Witten, E. Frank, L. Trigg, M. Hall G. Holmes, and S. J. Cunningham, Weka: Practical Machine Learning Tools and Techniques with Java Implementations, *Proceedings of the ICONIP/ANZIIS/ANNES'99 Workshop on Emerging Knowledge Engineering and Connectionist-Based Information Systems*, pp. 192-196, 1999.

[32] R. Werner Kristjanpoller, V. Kevin Michell, A stock market risk forecasting model through integration of switching regime, ANFIS and GARCH techniques, *Applied Soft Computing*, vol. 67, pp. 106-116, 2018.

[33] K. Udeh, D. W. Wanik, N. Bassill and E. Anagnostou, Time Series Modeling of Storm Outages with Weather Mesonet Data for Emergency Preparedness and Response, *2019 IEEE 10th Annual Ubiquitous Computing, Electronics & Mobile Communication Conference (UEMCON)*, New York City, NY, USA, pp. 0499-0505, 2019.

[34] A. Alodah, and O. Seidou, The adequacy of stochastically generated climate time series for water resources systems risk and performance assessment. *Stoch Environ Res Risk Assess*, vol. 33, pp. 253–269, 2019.

[35] A. Wibisono, J. Adibah, P. Mursanto, and M. S. Saputri, Improvement of Big Data Stream Mining Technique for Automatic Bone Age Assessment, *Proceedings of the 2019 ACM 3rd International Conference on Big Data Research*, pp. 119–123, 2019.

[36] P. Chouksey, and A. S. Chauhan, A Review of Weather Data Analytics using Big Data, *International Journal of Advanced Research in Computer and Communication Engineering*, vol. 6, no. 1, pp. 365-368, 2017.

[37] K. A. Ismail, M. A. Majid, J. M. Zain, and N. A. Abu Bakar, Big Data prediction framework for weather Temperature based on MapReduce algorithm, *2016 IEEE Conference on Open Systems (ICOS)*, Langkawi, pp. 13-17, 2016.

[38] K. A. Ismail, M. A. Majid, M. Fakherldin, and J. M. Zain, A Big Data Prediction Framework for Weather Forecast Using MapReduce Algorithm, Journal of Computational and Theoretical Nanoscience, vol. 23, no. 11, pp. 11138-11143(6), 2017.

[39] S. E. Haupt and B. Kosovic, Big Data and Machine Learning for Applied Weather Forecasts: Forecasting Solar Power for Utility Operations, *2015 IEEE Symposium Series on Computational Intelligence*, Cape Town, pp. 496-501, 2015.

[40] V. Dagade, M. Lagali, S. Avadhani, and P. Kalekar, Big Data Weather Analytics Using Hadoop, *International Journal of Emerging Technology in Computer Science & Electronics*, vol. 14, no. 2, pp. 847-851, 2015.

[41] T. Miyoshi, K. Kondo, and K. Terasaki, Big Ensemble Data Assimilation in Numerical Weather Prediction, in *Computer*, vol. 48, no. 11, pp. 15-21, 2015.

[42] J.N.K. Liu., Y. Hu, Y. He, P. W. Chan, and L. Lai, Deep Neural Network Modeling for Big Data Weather Forecasting. In: Pedrycz W., Chen SM. (eds) Information Granularity, Big Data, and Computational Intelligence. Studies in Big Data, vol 8, pp 389-408, Springer, Cham, 2015.

[43] J. Booz, W. Yu, G. Xu, D. Griffith, and N. Golmie, A Deep Learning-Based Weather Forecast System for Data Volume and Recency Analysis, *2019 International Conference on Computing, Networking and Communications (ICNC)*, Honolulu, HI, USA, pp. 697-701, 2019.

[44] S.-Y. Lin, C.-C. Chiang, J.-B. Li, Z.-S. Hung, and K.-M. Chao, Dynamic fine-tuning stacked auto-encoder neural network for weather forecast, *Future Generation Computer Systems*, vol. 89, Pages 446-454, 2018.

[45] F. J. Moreno, Sistema big data para mejorar los rendimientos agrícolas en Castilla y León, Degree dissertation, Udima, Madrid, Spain, 2019.

[46] P. Pyzel, Ampliación de un sistema de Big data para mejorar los rendimientos agrícolas con objetivo de realizar previsiones de necesidades de agua tratada en países con escasez de recursos hídricos, Degree dissertation, Udima, 2019.

[47] L. Breiman, Bagging predictors, *Machine Learning*, vol. 24, no. 2, pp.123-140, 1996.

# APPENDIX I – .ARFF files generated by the system

- A. Excerpt of a particular minable view created for "standard" analysis (file .arff).

```
@relation weather-project
@attribute Date date 'yyyy-MM-dd'
@attribute rainfall numeric
@attribute tmin numeric
@attribute tmax numeric
@data
2016-1-01,40.90490992906111,3.125,13.33111111111111
2016-2-01,34.753053538158774,5.157777777777778,18.84
2016-3-01,48.504665419434346,7.76046511627907,21.05625
2016-4-01,42.176677782541,12.59375,28.476829268292683
2016-5-01,?,14.482608695652175,29.78192771084337
2016-6-01,?,19.125555555555557,36.2038961038961
2016-7-01,?,20.276767676767676,36.09493670886076
2016-8-01,?,21.55056179775281,36.88076923076923
2016-9-01,?,16.78426966292135,32.72894736842105
2016-10-01,48.04021044733257,13.712903225806452,29.78170731707317
2016-11-01,?,7.062637362637362,21.71772151898734
2016-12-01,44.539838581248475,2.833707865168539,13.484210526315788
2017-1-01,32.95836866004329,1.4148936170212765,13.410975609756099
2017-2-01,36.37586159726386,1.1903225806451612,15.182894736842105
2017-3-01,40.60443010546419,6.790425531914893,20.07
2017-4-01,39.17010546939185,12.114285714285714,27.001785714285717
2017-5-01,?,14.576842105263157,31.349
2017-6-01,?,18.812222222222225,34.6
2017-7-01,?,22.743478260869566,38.703947368421055
2017-8-01,?,20.94123711340206,37.10253164556962
2017-9-01,?,18.993269230769233,34.8725
2017-10-01,0,14.519847328244273,27.939772727272725
2017-11-01,34.965075614664805,8.31359223300971,21.822093023255814
2017-12-01,40.16383020752389,5.935294117647059,19.084883720930232
```

- B. Excerpt of a particular minable view created for "neighbour-based" analysis (file .arff).

```
@relation weather-project
@attribute year numeric
@attribute month numeric
@attribute rainfall numeric
@attribute latitude numeric
@attribute longitude numeric
@attribute altitude numeric
@data
2016,1,3.258096538021482,325390,381950,686
2016,2,3.828641396489095,325390,381950,686
2016,3,5.399971020274537,325390,381950,686
2016,4,2.4849066497880004,325390,381950,686
2016,5,?,325390,381950,686
2016,6,?,325390,381950,686
2016,7,?,325390,381950,686
2016,8,?,325390,381950,686
2016,9,?,325390,381950,686
2016,10,?,325390,381950,686
2016,11,?,325390,381950,686
2016,12,3.349904087274605,325390,381950,686
2017,1,?,325390,381950,686
2017,2,?,325390,381950,686
2017,3,4.762173934797756,325390,381950,686
2017,4,4.269697449699962,325390,381950,686
2017,5,?,325390,381950,686
2017,6,?,325390,381950,686
2017,7,?,325390,381950,686
2017,8,?,325390,381950,686
2017,9,?,325390,381950,686
2017,10,?,325390,381950,686
2017,11,2.4849066497880004,325390,381950,686
2017,12,2.0149030205422647,325390,381950,686
2016,1,2.7950615780918397,321610,371490,677
```

- C. Excerpt of .ARFF test file.

```
@relation weather-project
@attribute year numeric
@attribute month numeric
@attribute rainfall numeric
@attribute latitude numeric
@attribute longitude numeric
@attribute altitude numeric
@data
2018,1,?,325390,381950,686
2018,2,?,325390,381950,686
2018,3,?,325390,381950,686
2018,4,?,325390,381950,686
2018,5,?,325390,381950,686
2018,6,?,325390,381950,686
2018,7,?,325390,381950,686
2018,8,?,325390,381950,686
2018,9,?,325390,381950,686
2018,10,?,325390,381950,686
2018,11,?,325390,381950,686
2018,12,?,325390,381950,686
```